\documentclass[final,10pt,1p,authoryear]{elsarticle}
\usepackage{cite}
\usepackage{amssymb}
\usepackage{amsmath}
\usepackage[figuresright]{rotating}
\usepackage{mathtools}
\usepackage{tabulary}
\usepackage{xparse}%
\usepackage{algpseudocode}
\usepackage{blindtext}
\usepackage[hidelinks,colorlinks=true,citecolor=blue]{hyperref}
\usepackage{makecell}
\usepackage{morewrites}
\usepackage{tikz}
\usetikzlibrary{arrows.meta,positioning,fit,shapes.geometric,calc}

\usepackage{wrapfig}
\usepackage{xkeyval}%
\usepackage{enumitem}
\usepackage{scalerel}
\usetikzlibrary{svg.path}
\definecolor{orcidlogocol}{HTML}{A6CE39}
\tikzset{orcidlogo/.pic={
\fill[orcidlogocol] svg{M256,128c0,70.7-57.3,128-128,128C57.3,256,0,198.7,0,128C0,57.3,57.3,0,128,0C198.7,0,256,57.3,256,128z};
\fill[white] svg{M86.3,186.2H70.9V79.1h15.4v48.4V186.2z}
                 svg{M108.9,79.1h41.6c39.6,0,57,28.3,57,53.6c0,27.5-21.5,53.6-56.8,53.6h-41.8V79.1z M124.3,172.4h24.5c34.9,0,42.9-26.5,42.9-39.7c0-21.5-13.7-39.7-43.7-39.7h-23.7V172.4z}
                 svg{M88.7,56.8c0,5.5-4.5,10.1-10.1,10.1c-5.6,0-10.1-4.6-10.1-10.1c0-5.6,4.5-10.1,10.1-10.1C84.2,46.7,88.7,51.3,88.7,56.8z};}}
\newcommand\orcidicon[1]{\href{https://orcid.org/#1}{\mbox{\scalerel*{\begin{tikzpicture}[yscale=-1,transform shape]
\pic{orcidlogo};\end{tikzpicture}}{|}}}}

\newwrite\authorbibfile%

\AtBeginDocument{%
  \immediate\openout\authorbibfile=\jobname.aub%
}%

\AtEndDocument{%
\immediate\closeout\authorbibfile
\InputIfFileExists{\jobname.aub}{}{}
}%

\makeatletter

\def\subsubsection{\@startsection{subsubsection}{3}{\z@}{0.5ex plus 0.1ex minus 0.1ex}{1ex plus 0.1ex}{\normalfont\normalsize\itshape}}
\def\BState{\State\hskip-\ALG@thistlm}

\define@key{authorbib}{scale}[1]{%
\def\AuthorbibKVMacroScale{#1}%
}

\define@key{authorbib}{wraplines}[10]{%
\def\AuthorbibKVMacroWraplines{#1}%
}

\define@key{authorbib}{imagewidth}[4cm]{%
\def\AuthorbibKVMacroImagewidth{#1}%
}

\define@key{authorbib}{overhang}[10pt]{%
\def\AuthorbibKVMacroOverhang{#1}%
}

\define@key{authorbib}{imagepos}[l]{%
\def\AuthorbibKVMacroImagepos{#1}%
}

\makeatother
\definecolor{tsgteal}{HTML}{2E8B8B}
\definecolor{tsgsalmon}{HTML}{D95F5A}
\definecolor{tsgnavy}{HTML}{1F3B5C}
\colorlet{goodcell}{tsgteal!18}     % "best" highlight (was goodcell)
\colorlet{badcell}{tsgsalmon!35}    % "poor" highlight (was badcell)
\presetkeys{authorbib}{imagepos=l,imagewidth=4cm,wraplines=15,overhang=20pt}{}

\newlength{\AuthorbibTopSkip}
\newlength{\AuthorbibBottomSkip}
\NewDocumentCommand{\authorbibliography}{+o+m+m+m}{%
  \IfNoValueTF{#1}{%
  }{%
    \setkeys{authorbib}{#1}%
    \immediate\write\authorbibfile{%
      \string\begin{wrapfigure}[\AuthorbibKVMacroWraplines]{\AuthorbibKVMacroImagepos}[\AuthorbibKVMacroOverhang]{\AuthorbibKVMacroImagewidth}^^J
        \string\includegraphics[scale=\AuthorbibKVMacroScale]{#2}^^J
        \string\end{wrapfigure}^^J
    }%
  }%
  \IfNoValueTF{#3}{%
    \typeout{Warning: No author name}%
  }{%
    \immediate\write\authorbibfile{%
      \unexpanded{\vspace{\AuthorbibTopSkip}}^^J
      \string\noindent\relax
      \unexpanded{{#3}\par}^^J
      \string\noindent\relax
      \unexpanded{#4}^^J%
      \unexpanded{\vspace{\AuthorbibBottomSkip}}^^J
      }%
  }%
}%
\usepackage{silence}
\usepackage{graphicx}
\usepackage{subcaption}
\usepackage{url}
\usepackage{multirow}
\usepackage{array}
\usepackage{color,soul}
\usepackage{colortbl}
\usepackage{pgf} % Required for pgfmath calculations

\definecolor{low}{rgb}{.1, 0.9, 1} %0.404, 0.628, .835
\definecolor{high}{rgb}{1, 0, 0.5} %.799, 0.26, 0.569

\usepackage[table,dvipsnames]{xcolor}
\usepackage{booktabs}
\usepackage{caption}
\usepackage{tablefootnote}
\usepackage{float}
\allowdisplaybreaks

\usepackage{pifont}

\usepackage{algorithm}

\usepackage{algpseudocode}
\algnewcommand\algorithmicinput{{Input:}}
\algnewcommand\Input{\item[\algorithmicinput]}
\algnewcommand\algorithmicoutput{{Output:}}
\algnewcommand\Output{\item[\algorithmicoutput]}
\algnewcommand\algorithmicdfne{{Define:}}
\algnewcommand\dfne{\item[\algorithmicdfne]}
\algnewcommand\algorithmicinitialize{{Initialize:}}
\algnewcommand\Initialize{\item[\algorithmicinitialize]}
\algnewcommand\algorithmicproc{{Procedure}}
\algnewcommand\proc{\item[\algorithmicproc]}
\algnewcommand\algorithmiceproc{{End Procedure}}
\algnewcommand\eproc{\item[\algorithmiceproc]}
\algnewcommand\algorithmicbefd{\underline{Before Deployment:}}
\algnewcommand\befd{\item[\algorithmicbefd]}
\algnewcommand\algorithmicaftd{\underline{After Deployment:}}
\algnewcommand\aftd{\item[\algorithmicaftd]}

\DeclareCaptionStyle{tabl}{labelformat=simple, labelsep=newline,font={scriptsize,sc}, justification=centering}
\DeclareCaptionStyle{figg}{labelformat=simple, labelsep=period,font=scriptsize, justification=centering}
\DeclareCaptionFormat{algor}{%
  \hrulefill\par\offinterlineskip\vskip1pt%
    {#1#2}#3\offinterlineskip\hrulefill}
\DeclareCaptionStyle{algori}{singlelinecheck=off,format=algor,labelsep=space}
\usepackage{geometry}
\usepackage{placeins}
\usepackage{tikz}
\usepackage{amsthm}
\theoremstyle{definition}

\usepackage{etoolbox}
\makeatletter
\def\ps@pprintTitle{%
   \let\@oddhead\@empty
   \let\@evenhead\@empty
   \let\@oddfoot\@empty
   \let\@evenfoot\@oddfoot
}
\makeatother





\begin{document}
\begin{frontmatter}
\title{Learning-based Probabilistic Load Forecasting with Post-hoc
and In-model Uncertainty\tnoteref{label0}}

\author[um,itu]{Sarah Al-Shareeda}
\author[itu]{Gulcihan Ozdemir\corref{cor1}}
\cortext[cor1]{Corresponding Author}
\author[ku]{Heung Seok Jeon}
\tnotetext[label0]{This work is supported by the Research Fund of the Istanbul Technical University Project Number: 47198. Authors Email addresses: salshareeda@um.edu.my and alshareeda@itu.edu.tr (Sarah Al-Shareeda), ozdemirg@itu.edu.tr (Gulcihan Ozdemir), and hsjeon@kku.ac.kr (Heung Seok Jeon).}

\affiliation[um]{{Computer System and Technology Department, Faculty of Computer Science and Information Technology, University of Malaya, Malaysia}}

\affiliation[itu]{{Informatics Institute, Istanbul Technical University, Turkey}}

\affiliation[ku]{Computer Engineering Department, Konkuk University, South Korea}

\begin{abstract}
Smart-building load forecasters are often trained offline on dense, multivariate, high-frequency data, but deployment may provide only hourly, feature-limited inputs. Missing features must then be reconstructed, and their errors can propagate through the model. If this input uncertainty is not reflected, prediction intervals may become miscalibrated, affecting demand-response scheduling. Our work examines where uncertainty should be placed once inference inputs are reconstructed. We develop a unified one-day-ahead probabilistic forecasting framework that aligns temporal resolution, reconstructs the unavailable inputs, and derives causal features, and we compare a modular post-hoc residual-quantile scheme with an integrated in-model quantile-learning scheme. The comparison uses three mid-scale Deep Learning (DL) backbones: recurrent, hybrid recurrent, and attention-based Temporal Fusion Transformer (TFT) models, under identical inputs, forecasting horizon, preprocessing rules, and training budgets. Results show that uncertainty placement is backbone-dependent. Integrated quantile learning is most reliable with the TFT, yielding 2.2-3.6\% MAPE and 28-83\,W RMSE on the labeled test window, while producing intervals about 5$\times$ narrower than the modular intervals at the closest-to-nominal coverage level. Diebold-Mariano tests support the TFT ranking and the mixed behavior of the recurrent backbones. A reconstruction-sensitivity test shows that reconstructed inputs increase the Quantile Score (QS) by 106\% while interval width remains nearly unchanged, indicating that the model does not automatically absorb reconstruction-induced uncertainty. Robustness checks against non-DL baselines and seasonal hold-out weeks support this ranking. Our results expose the limits of post-hoc residual quantiles when inference depends on reconstructed inputs.
\end{abstract}

\end{frontmatter}

\section{Introduction}\label{background}
Smart-grid and smart-building operation is seeing an increased reliance on forecasts to guide resource management and scheduling. With accurate forecasts, operators can efficiently coordinate Photovoltaic (PV) generation and battery storage, thereby reducing unnecessary operating costs \citet{arevalo2025smart,almihat2025role,el2024revolutionizing}. However, in practice, forecasting does not occur under ideal inference conditions. Deployment data may be sparse, low-resolution, and feature-limited due to sensor faults and communication disturbances. As a result, forecasting models trained with rich, multivariate, and high-resolution data may face asymmetric conditions at inference \citet{sarah2025lightweight}. Feature asymmetry requires missing inputs to be reconstructed prior to forecasting. Since reconstructed variables are proxy inputs and not true measurements, reconstruction error can propagate through the forecasting model \citet{11359042}. In probabilistic forecasting, the predicted distribution is then conditioned on such reconstructed inputs as if they were reliable. If this additional \emph{input uncertainty} is not learned or reflected, the resulting prediction intervals may become miscalibrated. How well a forecasting model absorbs this input uncertainty depends on where uncertainty is built into the forecasting process. Probabilistic forecasting methods address this design choice in modular and integrated ways. In the modular probabilistic scheme, a point forecaster is trained first and uncertainty is appended \textcolor{black}{post-hoc} using residual quantiles or error distributions \citet{khajeh2022applications,kaur2022energy,meisenbacher2022review}. In the integrated, \textcolor{black}{in-model} scheme, uncertainty is learned during training by embedding a pinball loss, a distributional likelihood, or Bayesian inference \citet{lucas2021probabilistic,brusaferri2022probabilistic,mahajan2024bayesian,ozdemir2024probabilistic}. \textcolor{black}{Although both designs are well established, they are rarely compared under reconstructed, feature-limited inference; which placement keeps intervals calibrated once the inputs are reconstructed is the open question this work examines.}

To answer this question, this paper develops a unified probabilistic forecasting framework for one-day-ahead smart-building active-power prediction under feature-asymmetric deployment. The framework reconstructs the unavailable inference-time inputs and evaluates three mid-scale Deep Learning (DL) backbones, namely Bidirectional Long Short-Term Memory (BiLSTM), Bidirectional Gated Recurrent Unit-LSTM (BiGRU-LSTM), and Temporal Fusion Transformer (TFT), under two uncertainty placements: a modular post-hoc residual-quantile scheme and an integrated in-model quantile-learning scheme. Our controlled setup isolates the effect of uncertainty placement on accuracy, calibration, sharpness, and operational behavior. As positioned in Fig.~\ref{fig:taxonomy}, we do not propose a new forecasting architecture; rather, we make the following contributions:
\begin{figure}[!htbp]
\centering
\includegraphics[width=\columnwidth]{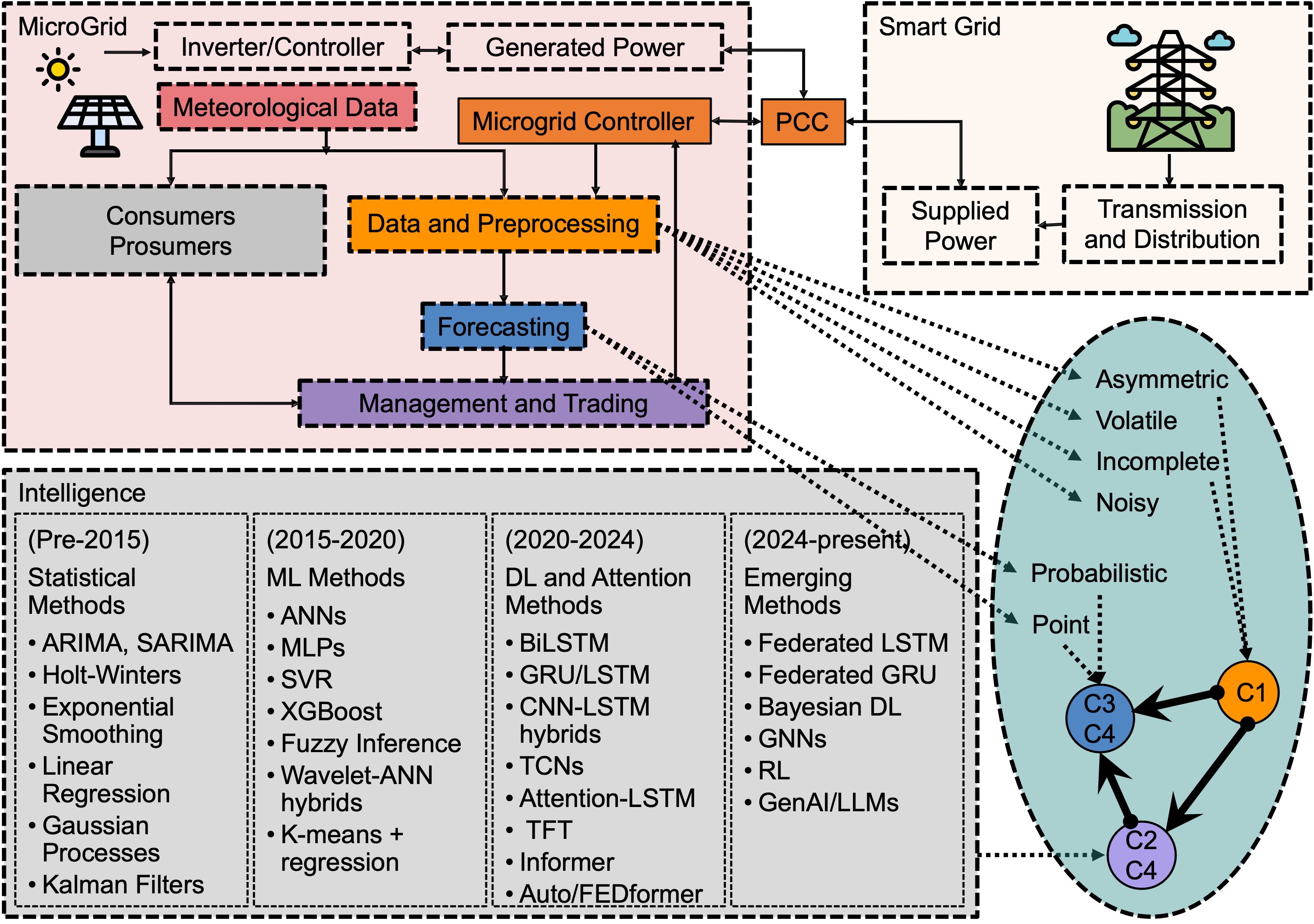}
\caption{Positioning of this work contributions.}
\label{fig:taxonomy}
\end{figure}
\begin{itemize}
\item [C1.] A unified one-day-ahead probabilistic forecasting framework is proposed for smart-building active-power prediction under feature-asymmetric deployment. The framework harmonizes the temporal resolution of the official 2025 Electric Energy Consumption Forecasting Competition dataset \citet{gomes_2024_14275645}, reconstructs unavailable inference-time features using validation-selected estimators, derives causal temporal features, and applies train-only normalization.

\item [C2.] A controlled uncertainty-placement benchmark is developed using three representative mid-scale DL backbones: BiLSTM, BiGRU-LSTM, and TFT. The benchmark compares a modular post-hoc residual-quantile scheme with an integrated in-model quantile-learning scheme under identical data splits, reconstructed inputs, forecasting horizon, recursive rollout, preprocessing rules, and training budget.

\item [C3.] The effect of reconstruction-induced input uncertainty on probabilistic forecast quality is assessed under feature-limited inference. The two uncertainty placements are evaluated using point-forecast accuracy, quantile loss, interval coverage, interval width, interval score, and statistical significance tests.

\item [C4.] The operational reliability of the uncertainty schemes is evaluated beyond aggregate error metrics. The analysis examines peak tracking, ramp-rate behavior, and demand-temperature loop consistency on the labeled test window, and further tests robustness against non-DL baselines, input-reconstruction choices, and seasonal hold-out periods.
\end{itemize}

The remainder of the paper is organized as follows. Section~\ref{lit} reviews related work, with particular emphasis on smart-grid data preprocessing, feature reconstruction, and uncertainty-aware load forecasting. Section~\ref{sec:framework} presents the unified framework, describing the forecasting models employed and the two uncertainty schemes. Section~\ref{result} outlines the experimental setup, evaluation metrics, and results. Section~\ref{conc} concludes with key findings and future directions.

\section{Literature Review}\label{lit}
Short-term load forecasting supports the reliable, cost-efficient operation of smart buildings and grids. Because the deployment setting is typically feature-asymmetric, forecast uncertainty becomes as important as point accuracy, especially when operating margins are tight. Our review focuses on preprocessing methods that clean, impute, align, or reconstruct smart-building data, and forecasting methods that estimate future demand as point values or probabilistic outputs.

\subsection{Smart\,Grid Data Challenges and Preprocessing}
Reliable load forecasting depends on accurate, complete, and temporally consistent inputs, but smart-building telemetry often violates these conditions, and preprocessing repairs such defects before forecasting. Sensor faults and communication dropouts create missing values, which data-quality methods address through imputation, denoising, feature fusion, and outlier removal; for example, \citet{Data_Purification} use a Reinforcement Learning (RL) agent to score data reliability before filtering low-quality entries, improving dispatch robustness under renewable-energy uncertainty (RL has been applied to decision-making under uncertainty in \citet{11106806,11054162,10978324}). Heterogeneous sampling rates cause temporal-resolution mismatch, and at deployment the model trained on dense multivariate data must infer from sparse, feature-limited inputs; alignment and reconstruction methods handle both, resampling heterogeneous data streams and estimating unavailable channels before forecasting \citet{sundararajan2022data,Multioutput}. Noisy or adversarial readings can bias the feature-target relationship, which robustness-oriented methods counter; the Deep Deterministic Policy Gradient (DDPG) framework of \citet{Robust_PV1}, for example, reweights multi-modal features to enhance PV forecasts. Consumption data are also privacy-sensitive, so raw measurements often cannot be centralized or freely shared across buildings; privacy-preserving methods use distributed measurements without exposing raw data, as in blockchain-based federated BiLSTM training for Electric Vehicle (EV) load \citet{Block_FeDL} and multi-party cryptographic computation for household forecasts \citet{PrivGrid}. \textcolor{black}{Preprocessing improves data quality, but the uncertainty it adds, especially when missing channels are reconstructed, is not carried into the forecast intervals. Reconstruction restores the expected input structure without recovering the real measurements, so the filled-in channels add no information beyond the signals available at inference. Our framework in Section~\ref{sec:framework} therefore tests the forecasts using the same reconstructed inputs the model would face in deployment, so we can measure how much the reconstruction affects interval calibration and sharpness instead of assuming it has no effect.}

\subsection{Uncertainty-unaware and Uncertainty-aware Forecasting}
Forecasting methods have advanced through several generations. Early statistical models, AutoRegressive Integrated Moving Average (ARIMA), Seasonal ARIMA (SARIMA), Holt-Winters, Gaussian Processes (GP), and Kalman Filters (KF), are computationally lightweight \citet{ozdemir2024probabilistic}, but their linear structure struggles with nonlinear demand and long temporal dependencies. Classical Machine Learning (ML) models loosen these limits; Back-Propagation NN (BPNN), Multi-Layer Perceptrons (MLP), Support Vector Regression (SVR), tree ensembles such as Gradient Boosting Machines (GBM) and Random Forests (RF), and wavelet-NN hybrids capture nonlinear structure \citet{ozdemir2024long}, but depend on manual feature engineering and poorly scale with the high dimension of smart-grid data. Deep Learning (DL) removes much of the burden; for example, LSTM and GRU networks learn long-range dependencies directly. Hybrid DL designs extend the capabilities; \citet{el2025enhancing} use ARIMA-LSTM to forecast EV-charging demand, and the LSTM-Transformer of \citet{pentsos2025hybrid} couples sequential memory with global attention at higher data and computation cost. More specialized AI methods have followed. \citet{zhang2025ultra} fine-tune a GPT variant for ultra-short-term PV forecasting, improving sparse-site accuracy at the cost of interpretability and data demand, and RL has been used to tie load prediction to asset-aware control \citet{zhang2024load}.

Most of these models use a single point forecast, which is not enough as occupant-driven variability, Heating, Ventilation and Air Conditioning (HVAC) cycling, and behind-the-meter PV generation, and external-forecast errors all make the load uncertain. \textcolor{black}{Probabilistic load forecasting presents an answer, by modeling uncertainty inside the network or appending it afterward. In-model methods learn the predictive interval during training; quantile-regression heads add an output layer that minimizes the pinball loss and returns a set of conditional quantiles directly, with no assumption about the shape of the distribution \citet{ZHANG2025303}, while Bayesian and distributional heads, Bayesian Neural Networks (BNN), Mixture Density Networks (MDN), and parametric likelihood layers, place a full predictive density on the target \citet{mahajan2024bayesian,brusaferri2022probabilistic,ozdemir2024probabilistic}. Post-hoc methods take a different route, leaving a trained point forecaster untouched and fitting a separate error model, empirical residual quantiles, Kernel Density Estimation (KDE), or conformal calibration, to its residuals \citet{khajeh2022applications,kaur2022energy,lucas2021probabilistic}. Post-hoc methods are model-agnostic, need only a held-out calibration set, and add little training cost, but the appended module never sees the forecaster's internal state, so its intervals lose calibration and sharpness once the inputs become noisy, incomplete, or distribution-shifted. Among post-hoc options, residual-quantile estimation is the most widely used. Related post-hoc interval methods occur across adjacent forecasting domains, including an LSTM-GRU backbone with KDE intervals for wave height \citet{wangying2023wave}, a decomposition-GRU with KDE intervals for offshore wind speed \citet{wang2024offshorewind}, and a Temporal Convolutional Network (TCN)-GRU-attention model with error-distribution analysis for multistep wave height \citet{wang2025waveheight}.}

\textcolor{black}{Although in-model and post-hoc uncertainty methods are well studied individually, their effect is rarely isolated under feature-asymmetric deployment, where models trained on rich, high-resolution data must forecast from sparse, feature-limited inputs. We compare modular uncertainty and integrated quantile learning using the same DL backbones, BiLSTM, BiGRU-LSTM, and TFT, within a fixed forecasting pipeline. Data, features, reconstruction, rollout, horizon, and training budget are held constant, so differences in accuracy, calibration, and sharpness reflect the uncertainty formulation. Our study therefore quantifies the cost of the post-hoc route rather than assuming its reliability.} Table~\ref{tab:lit_compare} summarizes representative studies and positions our work against them.

\begin{table*}[!htbp]
\centering
\caption{Representative studies across preprocessing and forecasting. The Uncertainty column reports how each study treats uncertainty: None for a point forecast, Post-hoc for intervals appended to a trained point model, In-model for uncertainty learned within the model, Mixed for surveys spanning several approaches, and N/A for preprocessing-only studies that produce no forecast.}
\label{tab:lit_compare}
\resizebox{\linewidth}{!}{
\begin{tabular}{lllp{9cm}}\hline
{Study} & {Domain} & {Uncertainty} & {Approach} \\\hline

\multicolumn{4}{l}{{Surveys / Reviews}}\\\hline
\citet{khajeh2022applications} & Smart grids & Post-hoc & Review of probabilistic forecasting; uncertainty estimation mainly residual-based \\
\citet{kaur2022energy} & Smart grids & Mixed & Survey of probabilistic deep learning for energy forecasting \\
\citet{meisenbacher2022review} & Time Series & Mixed & Automated time series pipelines; preprocessing-forecasting fragmentation \\\hline

\multicolumn{4}{l}{{Data preprocessing only}}\\\hline
\citet{Data_Purification} & Power dispatch & N/A & RL-based filtering to improve forecast input quality \\
\citet{Block_FeDL} & EV charging & N/A & Federated BiLSTM + blockchain for privacy \\
\citet{PrivGrid} & Household load & N/A & Cryptographic Multi-Party Computation (MPC) + LSTM; secure data sharing \\
\citet{Robust_PV1} & PV forecasting & N/A & RL-based robustness against adversarial noise \\\hline

\multicolumn{4}{l}{{Forecasting: point}}\\\hline
\citet{el2025enhancing} & EV charging & None & ARIMA + LSTM for ops optimization \\
\citet{pentsos2025hybrid} & Residential load & None & Hybrid LSTM + Transformer; compute-intensive \\
\citet{zhang2025ultra} & PV generation & None & GPT fine-tuning for ultra-short-term PV \\
\citet{zhang2024load} & Energy management & None & DL + RL; links load forecasts to battery aging \\
\citet{ozdemir2024long} & Long-term demand & None & Stepwise Linear Regression (SLR) vs. NN, Genetic Algorithm (GA), Differential Evolution (DE), Particle Swarm Optimization (PSO), Gaussian Process Regression (GPR) for long-term load \\\hline

\multicolumn{4}{l}{{Forecasting: probabilistic}}\\\hline
\citet{mahajan2024bayesian} & Building energy & In-model & Bayesian NNs for calibrated intervals \\
\textcolor{black}{\citet{brusaferri2022probabilistic}} & \textcolor{black}{Electric load} & \textcolor{black}{In-model} & \textcolor{black}{Bayesian MDN} \\
\citet{ozdemir2024probabilistic} & Distribution load & In-model & Probabilistic load via Normal/Weibull/Gamma/Lognormal fits \\
\textcolor{black}{\citet{ZHANG2025303}} & \textcolor{black}{Multi-energy load} & \textcolor{black}{In-model} & \textcolor{black}{Quantile-regression PatchTST; pinball-trained quantiles} \\
\citet{lucas2021probabilistic} & Buildings & Post-hoc & Building Energy Model (BEM) + Kernel Density Estimation (KDE) residuals for interval forecasts \\
\textcolor{black}{\citet{wangying2023wave}} & \textcolor{black}{Wave height} & \textcolor{black}{Post-hoc} & \textcolor{black}{LSTM-GRU point + post-hoc residual intervals} \\
\textcolor{black}{\citet{wang2024offshorewind}} & \textcolor{black}{Offshore wind} & \textcolor{black}{Post-hoc} & \textcolor{black}{Decomposition-GRU; post-hoc residual intervals} \\
\textcolor{black}{\citet{wang2025waveheight}} & \textcolor{black}{Wave height} & \textcolor{black}{Post-hoc} & \textcolor{black}{TCN-GRU-attention; post-hoc KDE intervals} \\
\hline

\multicolumn{4}{l}{{Hybrid preprocessing + forecasting}}\\\hline
\citet{Multioutput} & Smart-grid time series & In-model & Multi-output GP; joint imputation+forecast \\
\citet{MCSD_Transformer} & Smart-grid time series & None & Self-distillation Transformer; robust to missingness \\
\citet{sundararajan2022data} & PV + load & None & Data-quality-aware switching for drift/missingness \\
\citet{sarah2025lightweight} & Smart building/grid & None & Lightweight GRU-LSTM + preprocessing; handles gaps \\\hline

\multicolumn{4}{l}{{This study}}\\\hline
{Modular} & Smart building & Post-hoc & BiLSTM, BiGRU-LSTM, TFT; residual-based quantiles \\
{Integrated} & Smart building & In-model & BiLSTM, BiGRU-LSTM, TFT; pinball-trained quantiles \\\hline

\end{tabular}}
\end{table*}

\section{Modular and Integrated Probabilistic Forecasting Framework}\label{sec:framework}
\begin{figure*}[!htbp]
\centering
\includegraphics[width=\linewidth]{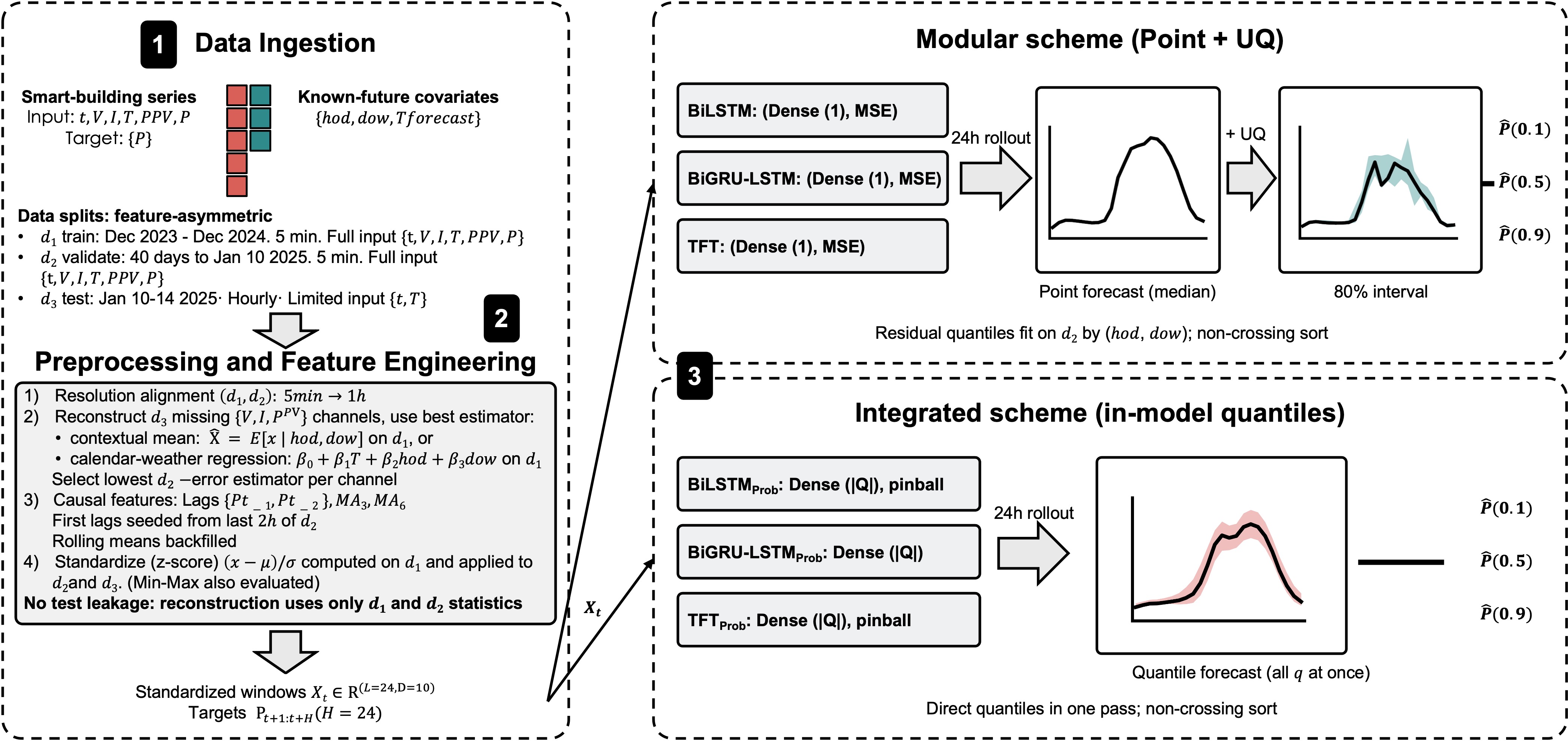}
\caption{Proposed three-stage framework for probabilistic load forecasting under feature-asymmetric deployment. Stages~1 and~2 ingest and align the data; Stage~3 compares the BiLSTM, BiGRU-LSTM, and TFT models under the modular and integrated schemes.}
\label{fig:blockdiagram}
\end{figure*}

Operational load forecasting has two requirements: predicting a building's day-ahead consumption and quantifying prediction uncertainty. In this section, we present the framework shown in Fig.~\ref{fig:blockdiagram} to address both under feature-asymmetric deployment, where inference-time inputs are sparser and lower resolution than the training data. The framework consists of three stages: 1)~smart-building data ingestion; 2)~preprocessing and feature engineering; and 3)~DL-based forecasting with BiLSTM, BiGRU-LSTM and TFT models. Stages~1 and~2 prepare and align the data, while Stage~3 performs the core evaluation by comparing the forecasting models under two uncertainty schemes: a {modular} scheme, where uncertainty is added after training a point forecaster, and an {integrated} scheme, where quantile forecasts are directly learned.

\subsection{Problem Formulation}\label{sec:problem}
We address one-day-ahead probabilistic forecasting of smart-building active power consumption under feature-asymmetric deployment. At each time step $t$, the objective is to predict the next $H$ hourly load values using the most recent $L$ historical observations and the known future inputs available over the prediction horizon, such as calendar variables and ambient temperature $T$. The forecasting sample is defined as
\begin{equation}
S_t=\left(X_t,\,Z_{t+1:t+H},\,P_{t+1:t+H}\right),
\label{eq:window}
\end{equation}
where
\begin{equation}
X_t=\{x_{t-L+1},\ldots,x_t\}\in\mathbb{R}^{L\times D}
\label{eq:input_window}
\end{equation}
is the historical input window with $D$ features per time step, $Z_{t+1:t+H}$ is the set of known future inputs over the forecasting horizon, and
\begin{equation}
P_{t+1:t+H}=\{P_{t+1},\ldots,P_{t+H}\}
\label{eq:target_window}
\end{equation}
is the target active-power trajectory. Both the input length $L$ and forecasting horizon $H$ are set to 24 hourly samples.

The goal is to estimate the future load trajectory $P_{t+1:t+H}$ and quantify its uncertainty. For each forecast step $h=1,\ldots,H$, the model predicts conditional quantiles
\begin{equation}
\widehat{P}^{(q)}_{t+h}=f_{\theta}^{(q)}\!\left(X_t,Z_{t+1:t+H}\right),\quad q\in{Q},
\label{eq:quantile_forecast}
\end{equation}
where ${Q}$ is the set of target quantile levels and $\theta$ denotes the trainable parameters. The conditioning window is advanced over the prediction horizon at inference time, as detailed in Section~\ref{sec:models}. The resulting intervals are evaluated by calibration, measuring whether they contain the true load at the expected rate, and sharpness, measuring whether they remain sufficiently narrow. The main challenge is the train-deployment asymmetry. During training, the model observes a rich, high-resolution feature set $F_{\text{train}}$, while inference uses a reduced, low-resolution set $F_{\text{test}}$. In particular, only timestamps and ambient temperature $T$ are available at inference, while the target load $P$ and several electrical channels are unobserved. The feature mismatch $F_{\text{test}}\neq F_{\text{train}}$ defines the central constraint of the problem. We aim to determine which uncertainty design is more reliable under this constraint: the modular or the integrated scheme.

\subsection{Stage~1: Smart-building Data Ingestion}\label{datasets}
The framework ingests a multivariate time series from a smart building:
\begin{equation}
{x}_t=\{V_t, I_t, T_t, P^{{PV}}_t, P_t\}, \quad t\in\mathbb{Z},
\end{equation}
where $V_t$ is voltage, $I_t$ is current, $T_t$ is ambient temperature, $P^{{PV}}_t$ is behind-the-meter PV generation, and $P_t$ is active power consumption, the forecasting target. From each timestamp $t$ we derive calendar features for hour-of-day ($hod$) and day-of-week ($dow$). Together with the provided temperature forecasts $\{hod_{t+h},dow_{t+h},T^{{forecast}}_{t+h}\}$, these are available in advance and treated as known-future inputs. The data are drawn from the official 2025 Electric Energy Consumption Forecasting Competition dataset~\citet{gomes_2024_14275645}, split as follows:
\begin{itemize}[leftmargin=1em]
\item Training set $d_1$: Dec\,01 2023 to Dec\,01 2024, sampled every 5\,min with the full feature set $\{t,V,I,T,P^{{PV}},P\}$.
\item Validation set $d_2$: Dec\,01 2024 to Jan\,10 2025 (40 consecutive days) at 5\,min resolution with the same features as $d_1$.
\item Test set $d_3$: Jan\,10 to Jan\,14 2025, sampled hourly with inputs restricted to $\{t,T\}$. The target $P$ is hidden at inference and revealed only for evaluation of Jan\,10-12, 2025.
\end{itemize}
These splits illustrate the train-test asymmetry of Section~\ref{sec:problem}: $d_1,d_2$ are multivariate and high-frequency; $d_3$ is sparse, hourly, and missing most features. \textcolor{black}{The validation and test windows fall in winter, so temperature and PV drop well below the training-year averages (Table~\ref{tab:dataset}). Active power and current are near-collinear ($P\approx VI$ at near-constant $V$), so the true current $I$ is target-equivalent and, like $P$, hidden at inference. Cross-building and multi-year generalization are out of the scope of this work; seasonal robustness within the year is assessed on held-out weeks as will be seen in Section~\ref{sec:results}.}
\begin{table}[!htbp]
\centering
\caption{\textcolor{black}{Dataset summary by split. $V, I, P^{{PV}}$ are observed in $d_1, d_2$ and reconstructed in $d_3$; $d_3$ inputs are $\{t, T\}$ and the target $P$ is hidden at inference (labeled days are Jan\,10-12, 2025).}}
\label{tab:dataset}
%\resizebox{\columnwidth}{!}{%
\begin{tabular}{lcccccc}
\toprule
Split & Period & Freq.\ & Hours & ${T}$\,($^\circ$C) & ${P^{{PV}}}$\,(W) & ${P}$\,(W) \\
\midrule
$d_1$ (train) & Dec\,01\,2023-Dec\,01\,2024 & 5\,min & 8{,}760 & 17.0 & 715 & $1909{\pm}1036$ \\
$d_2$ (val)   & Dec\,01\,2024-Jan\,10\,2025 & 5\,min & 960 & 12.5 & 258 & $2077{\pm}2028$ \\
$d_3$ (test)  & Jan\,10-14\,2025 & 1\,h & 120 & 12.5 & - & hidden \\
\bottomrule
\end{tabular}%}
\end{table}

\subsection{Stage~2: Preprocessing and Feature Engineering}\label{sec:stage2}
In Stage~2, the dense ($d_1,d_2$) and sparse ($d_3$) datasets are transformed into standardized supervised windows, constructed without using future or test-set information, as detailed below.
\subsubsection{Resolution Alignment}
Both $d_1$ and $d_2$ are sampled every 5\,min and $d_3$ every 1\,h. Therefore, we first downsample $d_1,d_2$ to 1\,h by a 12-point block average,
\begin{equation}
x'_k=\tfrac{1}{12}\textstyle\sum_{i=0}^{11}x_{12k+i},
\end{equation}
matching the hourly $d_3$ grid.

\subsubsection{Feature Reconstruction for $d_3$}
The deployment set $d_3$ is feature-deficient: while $d_1$ and $d_2$ contain $\{t,V,I,P^{PV},T,P\}$, only $\{t,T\}$ is available at inference time. To keep the input structure consistent, the missing channels $\{V,I,P^{PV}\}$ are reconstructed before forecasting. For each channel, two estimators are fitted on $d_1$, and the one with the lower validation error on $d_2$ is applied to $d_3$:
\begin{itemize}[leftmargin=1em]
\item Contextual mean filling: $\widehat{m}_{hod,dow}=\mathbb{E}[m\!\mid\!hod,dow]$, $\quad m\in\{V,I,P^{PV}\}$;
\item Calendar-weather regression: $\widehat{m}=\beta_0^{(m)}+\beta_1^{(m)}T+\beta_2^{(m)}hod+\beta_3^{(m)}dow$, $\quad m\in\{V,I,P^{PV}\}$, where $\{\beta_j^{(m)}\}_{j=0}^{3}$ are the regression coefficients estimated from $d_1$ for channel $m$. \end{itemize}
\textcolor{black}{The reconstruction uses only $d_1$ and $d_2$; the hidden $d_3$ targets are never used, so no test information leaks into the procedure. All fitting and selection are performed after aggregating $d_1$ and $d_2$ to the hourly inference grid, ensuring that training, validation, and deployment use the same temporal resolution. The reconstructed channels are used as proxy inputs to preserve a consistent feature structure between training, validation, and deployment. They are not recovered measurements or additional information beyond $(hod,dow,T)$. The practical effect of this reconstruction is examined through the sensitivity test in Section~\ref{sec:results}.}

\subsubsection{Causal Feature Engineering}
\textcolor{black}{To capture short-term load persistence, we add causal features from past active power: lags $\{P_{t-1},P_{t-2}\}$ and rolling means over 3 and 6\,h, $\{MA_3,MA_6\}$. On $d_1$ and $d_2$, they are computed from observed load. On $d_3$, where load is hidden at inference, the rollout is seeded from the last valid values of $d_2$ and the features are updated recursively from the model's predictions, so no hidden $d_3$ target is used.}

\subsubsection{Normalization and Supervised Windows}
With the channels reconstructed and the causal features derived, the per-timestep input is complete. It is standardized using $d_1$ statistics only, via standard scaling $x^*=(x-\mu)/\sigma$ with mean and standard deviation $(\mu,\sigma)$ estimated on $d_1$ and applied unchanged to $d_2$ and $d_3$, and assembled into the supervised windows of \eqref{eq:window} with per-timestep vector
\begin{equation}
\begin{aligned}
x_t=\big[&V_t,I_t,P^{PV}_t,T_t,hod_t,dow_t,P_{t-1},P_{t-2},MA_{3,t},MA_{6,t}\big],
\end{aligned}
\end{equation}
thus $D=10$, where $(V,I,P^{PV})$ are observed in $d_1,d_2$ and reconstructed in $d_3$.

\subsection{Stage~3: Forecasting with Modular and Integrated Schemes}\label{sec:models}
\begin{figure*}[!htbp]
\centering
\begin{subfigure}[t]{0.48\textwidth}\centering
    \includegraphics[width=\linewidth]{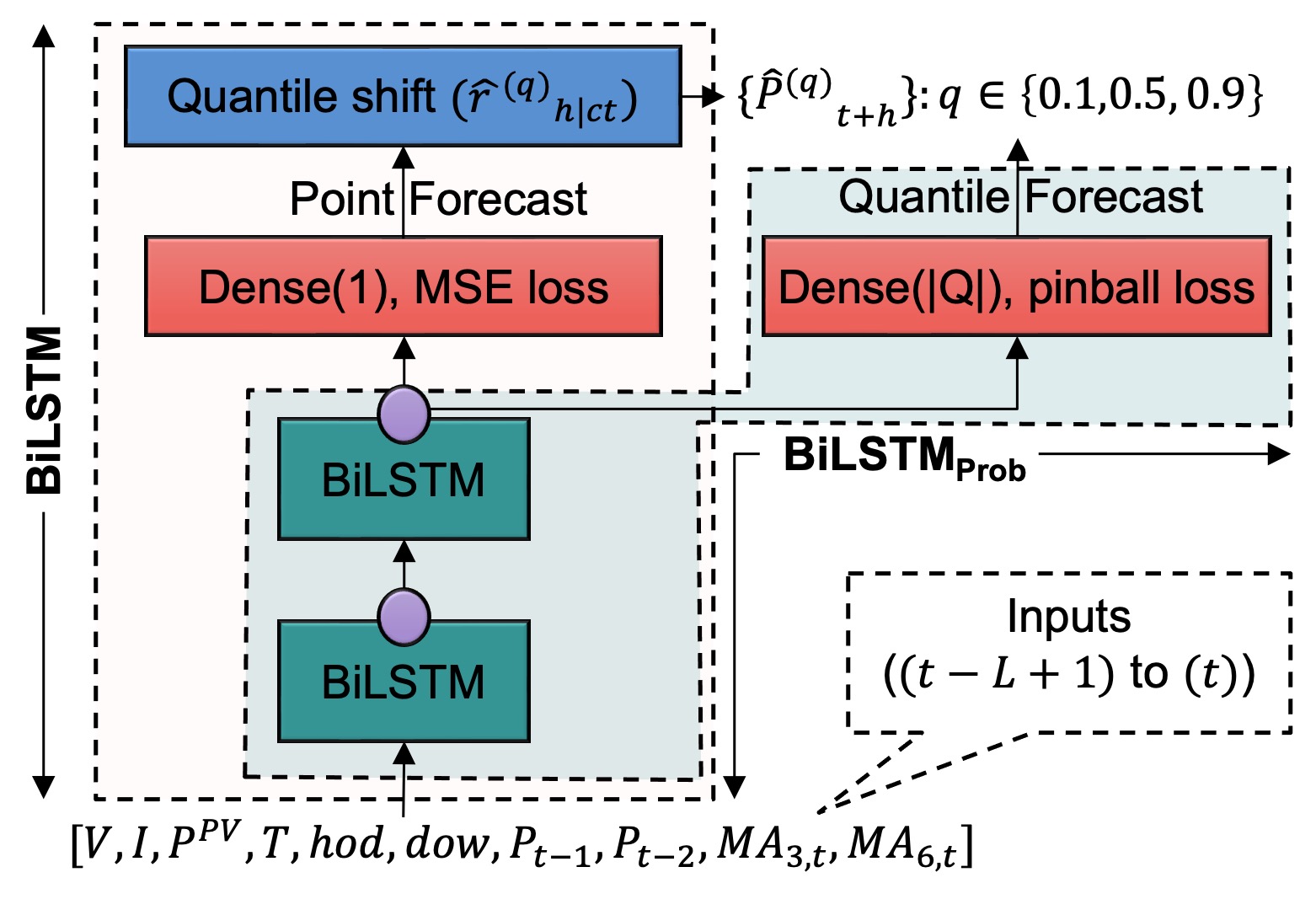}\caption{BiLSTM~/~BiLSTM$_{Prob}$.}\label{fig:bis}
\end{subfigure}\hspace{.1em}
\begin{subfigure}[t]{0.48\textwidth}\centering
    \includegraphics[width=\linewidth]{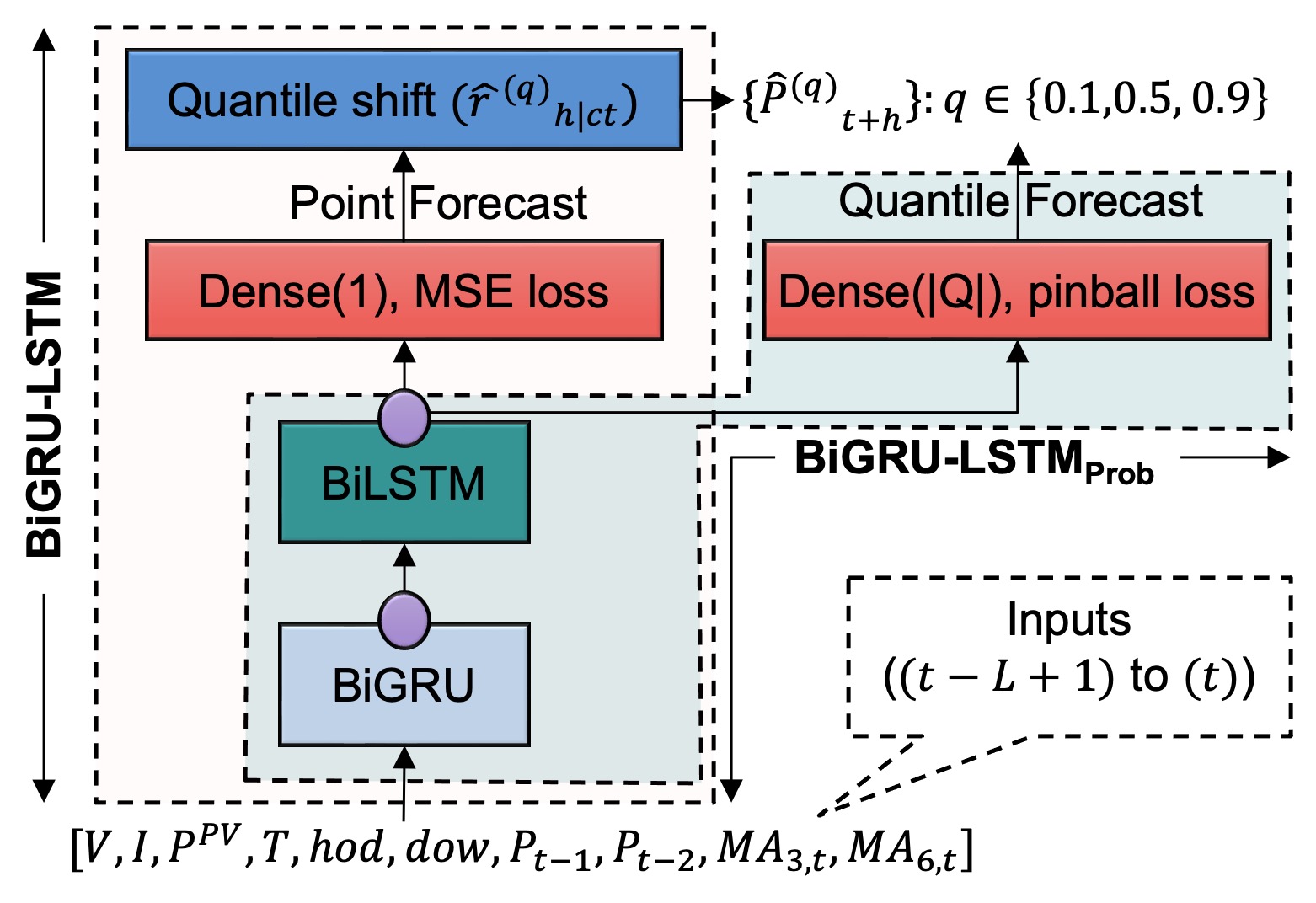}\caption{BiGRU-LSTM~/~BiGRU-LSTM$_{Prob}$.}\label{fig:big}
\end{subfigure}
  
\begin{subfigure}[t]{\textwidth}\centering
    \includegraphics[width=\linewidth]{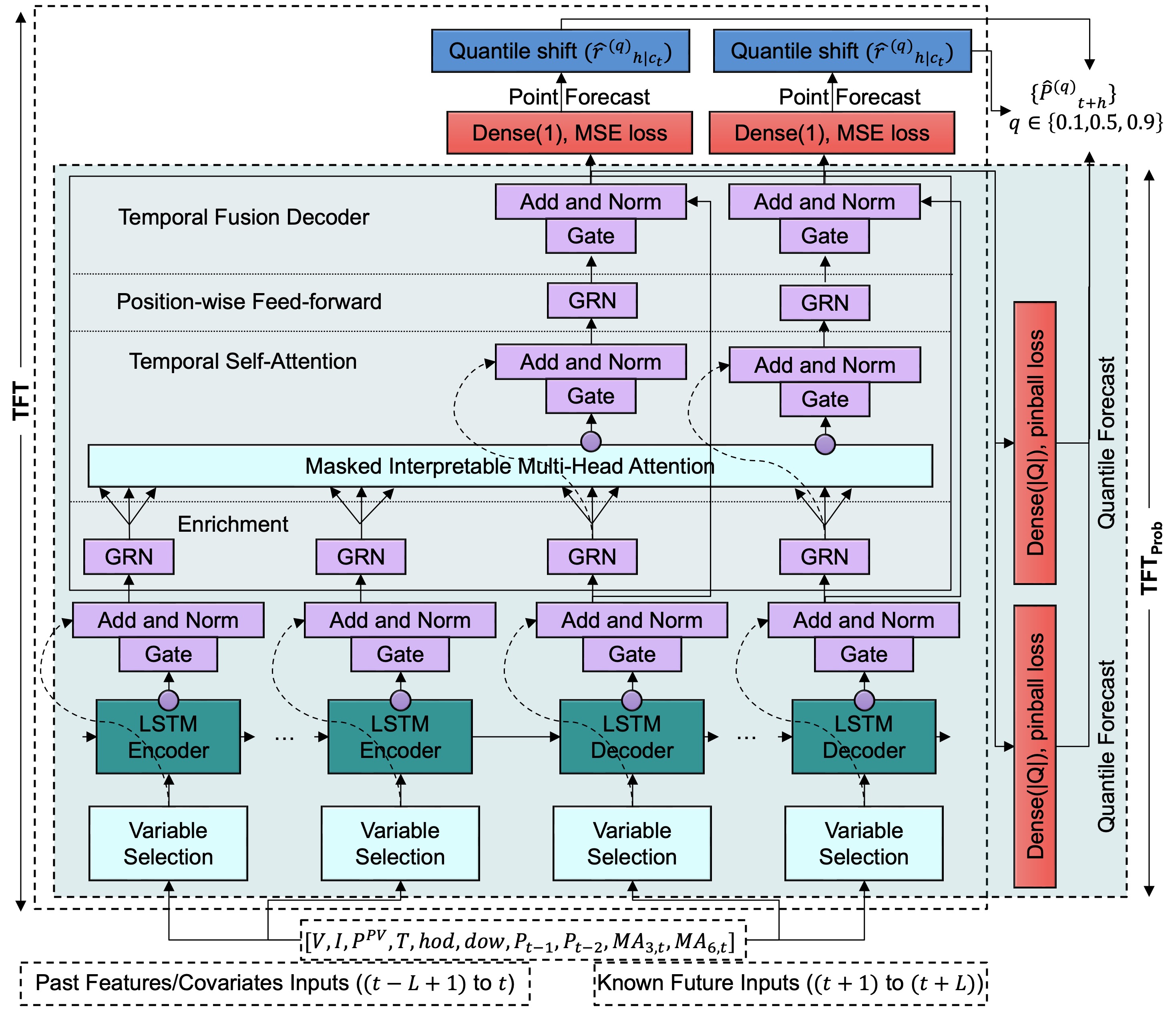}\caption{TFT~/~TFT$_{Prob}$.}\label{fig:tft}
\end{subfigure}
\caption{The three DL architectures under the two uncertainty schemes.}\label{fig:models2schemes}
\end{figure*}

Stage~3 compares two uncertainty schemes on the standardized inputs from Stage~2. We do not introduce new architectures but isolate the effect of how uncertainty is incorporated; both schemes share the input window and horizon of \eqref{eq:window}, data, features, and training, so performance differences arise only from the uncertainty formulation. We use three mid-scale models of increasing complexity, a recurrent model (BiLSTM), a hybrid recurrent model (BiGRU-LSTM), and an attention-based model (TFT), all established in smart-building and grid forecasting:
\begin{itemize}[leftmargin=1em]
\item {BiLSTM~/~BiLSTM$_\text{Prob}$} (Fig.~\ref{fig:models2schemes}\subref{fig:bis}): Two bidirectional LSTM layers encode the length-$L$ input window, with forward and backward hidden states concatenated, pooled, and regularized by dropout; reading both directions captures local context under noisy/imputed inputs. At each time step $t$, the LSTM cell maps the input $x_t$ and previous state $(h_{t-1},c_{t-1})$ to a new state through input ($i_t$), forget ($f_t$), and output ($o_t$) gates,
\begin{equation}
\begin{aligned}
i_t&=\sigma(w_i[h_{t-1},x_t]+b_i), & f_t&=\sigma(w_f[h_{t-1},x_t]+b_f),\\
o_t&=\sigma(w_o[h_{t-1},x_t]+b_o), & \tilde{c}_t&=\tanh(w_c[h_{t-1},x_t]+b_c),\\
c_t&=f_t\odot c_{t-1}+i_t\odot\tilde{c}_t, & h_t&=o_t\odot\tanh(c_t),
\end{aligned}
\end{equation}
where $\sigma(\cdot)$ is the logistic sigmoid, $\odot$ element-wise multiplication, $[\cdot,\cdot]$ concatenation, and $w,b$ the gate weights and biases.
\item {BiGRU-LSTM~/~BiGRU-LSTM$_\text{Prob}$} (Fig.~\ref{fig:models2schemes}\subref{fig:big}): A bidirectional GRU extracts short-term fluctuations and a unidirectional LSTM preserves slower cycles such as HVAC patterns, combining GRU efficiency with LSTM memory. The GRU cell uses update ($z_t$) and reset ($r_t$) gates,
\begin{equation}
\begin{aligned}
z_t&=\sigma(w_z[h_{t-1},x_t]+b_z), & r_t&=\sigma(w_r[h_{t-1},x_t]+b_r),\\
\tilde{h}_t&=\tanh(w_h[r_t\odot h_{t-1},x_t]+b_h), & h_t&=(1-z_t)\odot h_{t-1}+z_t\odot\tilde{h}_t.
\end{aligned}
\end{equation}
\item {TFT~/~TFT$_\text{Prob}$} (Fig.~\ref{fig:models2schemes}\subref{fig:tft}): To reweight features adaptively and model multi-scale dependencies, the TFT integrates Variable Selection Networks (VSNs), Gated Residual Networks (GRNs), local LSTMs, and multi-head self-attention in a fusion decoder. Its variable processing uses the GRN with Gated Linear Unit (GLU) and Exponential Linear Unit (ELU)
\begin{equation}
{GRN}(a)={LayerNorm}\!\big(a+{GLU}(w_2\,{ELU}(w_1 a))\big),
\end{equation}
combined with an LSTM encoder-decoder for local processing and interpretable multi-head attention for long-range dependencies, following the standard formulation~\citet{lim2021tft}. \textcolor{black}{The variable selection in the TFT, which emphasizes informative inputs and down-weights uninformative ones, parallels data-driven neural surrogate models that approximate expensive engineering simulations from limited or partially observed data using learned representations and adaptive feature weighting \citet{zhangimage2023,wangpore2024,zhang2022ijcm}. This shared ability to reweight inputs makes an attention-based backbone such as the TFT well suited to feature-asymmetric deployment, where some inputs are reconstructed or information-limited.}
\end{itemize}

The three models span complementary inductive biases: bidirectional recurrence for local context (BiLSTM), an added LSTM stage for longer cycles (BiGRU-LSTM), and attention with variable selection that spans the window directly, not only through recurrence (TFT). Hence, the modular-versus-integrated comparison holds across architectures rather than being confounded with a single backbone.

\subsubsection{Modular Scheme}
In the modular scheme, each model is a point forecaster with a Dense(1) head that predicts $P_{t+1}$ from $X_t$ under the Mean Squared Error (MSE) loss:
\begin{equation}
L_{{MSE}}(\theta)=\frac{1}{|d_1|}\sum_{t\in d_1}\big(P_{t+1}-g_\theta(\phi_\theta(X_t))\big)^2,
\label{eq:mse}
\end{equation}
where $\theta$ are the trainable parameters, $\phi_\theta(\cdot)$ the encoder, and $g_\theta(\cdot)$ the output head. Forecasts are produced recursively; at horizon step $h$, the prediction $\widehat{P}_{t+h}$ is inserted back into the lag and rolling features, and the updated input $X_{t+h}$ generates the next step. The rollout is seeded from the last $L$ hours of $d_2$, mimicking deployment. Uncertainty is then appended. At each $h$, the residual is
\begin{equation}\label{resid}
r_{t+h}=P_{t+h}-\widehat{P}^{{point}}_{t+h\mid t+h-1}.
\end{equation}
Residuals are grouped by horizon $h$ and context $c$. \textcolor{black}{The context $c_t=(hod_t,dow_t)$ is the calendar bucket of timestep $t$, and residuals are pooled within each bucket.} For each group, quantiles are estimated on $d_2$:
\begin{equation}
\widehat{r}^{(q)}_{h \mid c}=\operatorname{Quantile}_q\!\left(\left\{\, r_{t+h}\,:\,c_t=c,\ t\in d_2\right\}\right),
\label{eq:resid_quantile_empirical}
\end{equation}
with sparse cases smoothed by KDE. At inference, the contextual quantiles adjust the point forecast:
\begin{equation}
\widehat{P}^{(q)}_{t+h}=\widehat{P}^{{point}}_{t+h \mid t+h-1}+\widehat{r}^{(q)}_{h \mid c_t},\quad q\in{Q},
\label{eq:scheme1}
\end{equation}
and a non-crossing adjustment \textcolor{black}{(a per-step sort of the predicted quantiles)} enforces ordered intervals. Such augmentation is lightweight, requiring only contextual statistics in addition to the point forecaster. \textcolor{black}{We include it as a deliberate post-hoc baseline rather than a recommended method, to isolate the value of integrated quantile training.} \textcolor{black}{Algorithm~\ref{alg:scheme1} summarizes the procedure.}

\begin{algorithm}[H]
\caption{Modular residual-quantile forecasting}
\label{alg:scheme1}
\begin{algorithmic}[1]
\Input Training set $d_1$, validation set $d_2$, window $L{=}24$, horizon $H{=}24$, quantile set ${Q}$
\Output Quantile forecasts $\{\widehat{P}^{(q)}_{t+h}\}_{q\in{Q}}$ for $h=1,\dots,H$
\State Train the point model $g_\theta$ on $d_1$ with the MSE loss \eqref{eq:mse}.
\State Roll out recursively over $H{=}24$\,h, feeding each prediction back into the lag and rolling features.
\State On $d_2$, compute residuals \eqref{resid}, group by $(h,c)$, estimate quantiles \eqref{eq:resid_quantile_empirical}, and KDE-smooth sparse buckets.
\State At inference, add the contextual residual quantiles to the point forecast \eqref{eq:scheme1}; sort the quantiles to enforce non-crossing.
\end{algorithmic}
\end{algorithm}

\subsubsection{Integrated Scheme}
The integrated scheme embeds uncertainty directly in the forecaster. A Dense($|{Q}|$) head outputs all target quantiles simultaneously, trained with the quantile (pinball) loss:
\begin{equation}\label{pinball}
L_{Q}(\theta)=\frac{1}{|d_1|}\sum_{t\in d_1}\sum_{q\in{Q}}\rho_q\!\Big(P_{t+1}-g_\theta(\phi_\theta(X_t))^{(q)}\Big),
\end{equation}
where $\rho_q(u)=\max\{q\,u,(q-1)u\}$ asymmetrically penalizes under- and over-estimation. At inference, the model emits $\{\widehat{P}^{(q)}_{t+h}\}_{q\in{Q}}$ directly and rolls forward recursively over the $H{=}24$\,h horizon; the median forecast $\widehat{P}^{(0.5)}_{t+h}$ is fed back to update the lag and rolling features, as the point forecast is in the modular scheme. An optional non-crossing sort keeps the intervals ordered. Producing the full quantile set in one pass, the integrated scheme models the trajectory and its uncertainty jointly, without a separate residual step. \textcolor{black}{Algorithm~\ref{alg:scheme2} summarizes the procedure.}
\begin{algorithm}[H]
\caption{Integrated quantile forecasting}
\label{alg:scheme2}
\begin{algorithmic}[1]
\Input Training set $d_1$, window $L{=}24$, horizon $H{=}24$, quantile set ${Q}$
\Output Quantile forecasts $\{\widehat{P}^{(q)}_{t+h}\}_{q\in{Q}}$ for $h=1,\dots,H$
\State Train the model with a Dense($|{Q}|$) head minimizing the pinball loss \eqref{pinball}.
\State Roll out recursively over $H{=}24$\,h, feeding back the median $\widehat{P}^{(0.5)}$ to update lag and rolling features.
\State Output the quantiles directly; sort them to enforce non-crossing.
\end{algorithmic}
\end{algorithm}

Both schemes share recursive rollout and feature updates and use comparable parameter scales and training budgets, so any performance gap reflects the uncertainty formulation, not the rollout or model capacity. Next, we evaluate which scheme is more accurate and reliable under these asymmetric training-testing conditions.

\section{Simulations and Analysis}\label{result}
In this section, we present the experimental setup, training configuration, and results under the asymmetric train-test conditions. All experiments use the official competition dataset~\citet{gomes_2024_14275645}, which mirrors a deployment in which the full feature set is unavailable at inference. As in Section~\ref{datasets}, the data are partitioned into $d_1$ (training), $d_2$ (validation), and $d_3$ (test). The test set spans five consecutive days, Jan\,10-14, 2025, at 1\,h resolution, and the task is one day ahead, $H=24$. \textcolor{black}{Since the models predict one hour at a time, the $H=24$ horizon is generated by a recursive rollout; each prediction is reinserted into the input window to forecast the next hour until the full day is obtained.} Per the competition rules, only timestamped ambient temperature $T$ is available at inference, and the target $P$ is released later only for the labeled window (Jan\,10-12, 2025). The missing features $\{V, I, P^{PV}\}$ are reconstructed as in Stage~2. All variables are standardized with $d_1$ statistics, and outputs are inverse-transformed to Watts before evaluation.

All models are trained with Adam at batch size 32, with an initial Learning Rate (LR) of $10^{-3}$ for the recurrent models and $3\times10^{-4}$ for the TFT. Training stops early once the validation loss stops improving (patience 5 epochs for the recurrent models, 10 for the TFT), restoring the best weights. For the recurrent models, a scheduler halves the LR after 3 stagnant epochs, down to $10^{-6}$. Hyperparameters, in particular hidden size and dropout, are selected \textcolor{black}{by grid search} on $d_2$; \textcolor{black}{the per-model configurations are listed in Table~\ref{tab:tfthp}}. Architectures, training budgets, and preprocessing are identical across schemes. Experiments run on Google Colab (Python~3, Google Compute Engine backend) on a single NVIDIA A100 GPU with 40\,GB VRAM and 83.5\,GB RAM.
\begin{table}[!htbp]
\centering
\caption{\textcolor{black}{Model configurations.}}\label{tab:tfthp}
\begin{tabular}{lccc}\toprule
Setting & BiLSTM & BiGRU-LSTM & TFT \\\midrule
Input / output window & \multicolumn{3}{c}{24\,h / 24\,h} \\
Batch size & \multicolumn{3}{c}{32} \\
Quantiles & \multicolumn{3}{c}{$\{0.1,0.5,0.9\}$} \\
\midrule
Core architecture & 2$\times$BiLSTM & BiGRU + LSTM & 2$\times$LSTM + attention \\
Hidden units & \multicolumn{2}{c}{256,128} & 128 \\
Attention heads & \multicolumn{2}{c}{-} & 4 \\
Dropout & \multicolumn{2}{c}{0.3} & 0.2 \\
Optimizer / LR & \multicolumn{2}{c}{Adam / $10^{-3}$} & Adam / $3\times10^{-4}$ \\
Max epochs (patience) & \multicolumn{2}{c}{50 (5)} & 60 (10) \\
Total parameters & $\approx$1.2\,M & $\approx$0.74\,M & $\approx$1.2\,M\\\bottomrule
\end{tabular}
\end{table}
\subsection{Evaluation Metrics}
\textcolor{black}{Both schemes are assessed with a unified metric set spanning four aspects: point accuracy, probabilistic quality, interval calibration and sharpness, and efficiency~\citet{sarah2025lightweight}.}
\subsubsection{Point Accuracy}
\textcolor{black}{The prediction $\widehat{P}_k$ at step $k=1,\dots,N$ is the point forecast $\widehat{P}^{{point}}_k$ for the modular scheme and the median $\widehat{P}^{(0.5)}_k$ for the integrated scheme. It is scored against the observed load $P_k$ by the Root MSE (RMSE), Mean Absolute Error (MAE), Mean Absolute Percentage Error (MAPE), and a normalized accuracy score:}
\textcolor{black}{\begin{equation}
{RMSE}\,({W}) = \sqrt{\tfrac{1}{N}\textstyle\sum_{k=1}^N (\widehat{P}_k - P_k)^2},
\end{equation}}
\textcolor{black}{\begin{equation}
{MAE}\,({W}) = \tfrac{1}{N}\textstyle\sum_{k=1}^N |\widehat{P}_k - P_k|,
\end{equation}}
\textcolor{black}{\begin{equation}
{MAPE}\,(\%) = \tfrac{100}{N}\textstyle\sum_{k=1}^N \tfrac{|\widehat{P}_k - P_k|}{|P_k| + \varepsilon},
\end{equation}}
\textcolor{black}{\begin{equation}
{Accuracy}\,(\%) = 100\left(1 - \frac{\tfrac{1}{N}\sum_{k=1}^N |\widehat{P}_k - P_k|}{\max(P) - \min(P)}\right),
\end{equation}}
\textcolor{black}{with $\varepsilon$ guarding against division by zero.}
\subsubsection{Probabilistic Quality}
\textcolor{black}{The Quantile Score (QS) is the pinball loss averaged over ${Q}=\{0.1,0.5,0.9\}$:
\begin{equation}
{QS}\,({W}) = \tfrac{1}{|{Q}|}\tfrac{1}{N}\sum_{q\in{Q}}\sum_{k=1}^N \rho_q\!\big(P_k - \widehat{P}^{(q)}_k\big),
\end{equation}
where $\rho_q$ penalizes under- and over-prediction asymmetrically. Lower QS rewards sharper, statistically consistent forecasts; evaluated at the three quantiles, it is a discrete approximation of the Continuous Ranked Probability Score (CRPS).}
\subsubsection{Interval Calibration and Sharpness}
\textcolor{black}{At each step $k$, the $0.1$ and $0.9$ quantiles yield an 80\% interval $[Lo_k,Up_k]=[\widehat{P}^{(0.1)}_k,\widehat{P}^{(0.9)}_k]$, scored by the Prediction Interval Coverage Probability (PICP), the Mean Prediction Interval Width (MPIW), and the Mean Interval Score (MIS):
\begin{equation}
{PICP}_{80} = \tfrac{1}{N}\sum_{k=1}^N \mathbb{I}\{Lo_k \le P_k \le Up_k\},
\end{equation}
\begin{equation}
{MPIW}_{80}\,(W)= \tfrac{1}{N}\sum_{k=1}^N (Up_k - Lo_k),
\end{equation}
\begin{equation}
\begin{aligned}
{MIS}_{80}\,(W)= \tfrac{1}{N}\sum_{k=1}^N \big[&(Up_k - Lo_k)+\tfrac{2}{\alpha}(Lo_k - P_k)\,\mathbb{I}\{P_k < Lo_k\}+\tfrac{2}{\alpha}(P_k-Up_k)\,\mathbb{I}\{P_k > Up_k\}\big].
\end{aligned}
\end{equation}
${PICP}_{80}$ should approach $0.8$; lower values indicate under-coverage, higher ones mean over-coverage. ${MPIW}_{80}$~(W) measures sharpness, smaller is better. ${MIS}_{80}$~(W), with $\alpha=0.2$, is the mean Winkler score, penalizing both excess width and missed observations, weighing calibration against sharpness.}
\subsubsection{Model Computational Efficiency}
\textcolor{black}{Efficiency is reported as inference latency (sec) per 24\,h forecast.}

\subsection{Results and Discussion}\label{sec:results}
\textcolor{black}{Our experiments answer the question posed in Section~\ref{sec:problem}: under feature-asymmetric deployment, which uncertainty design yields more accurate and reliable forecasts, the modular or the integrated scheme, and does it depend on the backbone? As the active load for Jan\,13-14, 2025 is unreleased, evaluation covers Jan\,10-12. Throughout, \emph{point} denotes the point forecast of the modular pipeline before uncertainty augmentation, \emph{modular} its residual-quantile output, and \emph{integrated} the quantile-trained probabilistic model. We first assess forecast quality and its statistical significance. Next is the operational behavior. Finally, we analyze robustness against non-deep baselines, to the input reconstruction, and across seasons.}

\subsubsection{Forecast Quality and Significance}
\paragraph{Point-Forecast Accuracy}
\textcolor{black}{Table~\ref{tab:point_all} and Table~\ref{tab:perday_point} show that the scheme effect depends on the backbone. For TFT, integrated training is decisively better, 2.7\% MAPE and 99.0\% accuracy, with per-day RMSE of 28-83\,W. For the recurrent models the two schemes perform similarly; BiGRU-LSTM attains its lowest MAPE there. Across backbones, TFT leads by a wide margin while the recurrent models stay at 43-55\% MAPE for BiGRU-LSTM and 60-70\% for BiLSTM.}
\begin{table}[!htbp]
\centering
\caption{Point-forecast accuracy across models (W, \%). \colorbox{goodcell}{Teal} marks the best method within each model. \textcolor{black}{Values are means over the three labeled days (Jan\,10-12).}}
\label{tab:point_all}
\begin{tabular}{llcccc}
\toprule
Model & Method & RMSE$\downarrow$ & MAE$\downarrow$ & MAPE$\downarrow$ & Accuracy$\uparrow$ \\
\midrule
\multirow{4}{*}{BiLSTM} 
 & Point      & \cellcolor{goodcell}944.1 & 747.5 & 62.2 & 77.2 \\
 & Modular    & 1194.4 & 948.0 & 69.8 & 69.8 \\
 & Integrated & 1079.8 & \cellcolor{goodcell}740.1 & \cellcolor{goodcell}60.2 & \cellcolor{goodcell}77.7 \\
 & Average    & 1072.8 & 811.9 & 64.1 & 74.9 \\
\midrule
\multirow{4}{*}{BiGRU-LSTM} 
 & Point      & 814.8 & \cellcolor{goodcell}553.0 & 44.2 & \cellcolor{goodcell}83.7 \\
 & Modular    & \cellcolor{goodcell}812.2 & 605.4 & \cellcolor{goodcell}43.2 & 81.6 \\
 & Integrated & 923.4 & 651.8 & 55.1 & 80.5 \\
 & Average    & 850.1 & 603.4 & 47.5 & 81.9 \\
\midrule
\multirow{4}{*}{TFT} 
 & Point      & 227.0 & 181.0 & 14.4 & 94.9 \\
 & Modular    & 224.1 & 165.8 & 12.0 & 95.4 \\
 & Integrated & \cellcolor{goodcell}48.4 & \cellcolor{goodcell}35.7 & \cellcolor{goodcell}2.7 & \cellcolor{goodcell}99.0 \\
 & Average    & 166.5 & 127.5 & 9.7 & 96.4 \\
\bottomrule
\end{tabular}
\end{table}
\begin{table}[!htbp]
\centering
\caption{\textcolor{black}{Per-day point accuracy on the labeled days.}}
\label{tab:perday_point}
\begin{tabular}{lllcccc}
\toprule
Model & Day & Method & RMSE$\downarrow$ & MAE$\downarrow$ & MAPE$\downarrow$ & Acc$\uparrow$ \\
\midrule
\multirow{9}{*}{BiLSTM}
 & \multirow{3}{*}{Jan\,10} & Point  & 902.2 & 739.3 & 63.6 & 76.5 \\
 & & Modular & 1352.0 & 1110.3 & 75.5 & 64.7 \\
 & & Integrated & 996.6 & 644.6 & 57.3 & 79.5 \\
\cmidrule{2-7}
 & \multirow{3}{*}{Jan\,11} & Point  & 1012.7 & 772.2 & 59.5 & 70.5 \\
 & & Modular & 1420.7 & 1092.4 & 69.8 & 58.3 \\
 & & Integrated & 1097.2 & 765.1 & 60.5 & 70.8 \\
\cmidrule{2-7}
 & \multirow{3}{*}{Jan\,12} & Point  & 917.4 & 731.0 & 63.6 & 84.6 \\
 & & Modular & 810.6 & 641.5 & 64.2 & 86.5 \\
 & & Integrated & 1145.6 & 810.6 & 63.0 & 82.9 \\
\midrule
\multirow{9}{*}{BiGRU-LSTM}
 & \multirow{3}{*}{Jan\,10} & Point  & 677.4 & 433.2 & 41.2 & 86.2 \\
 & & Modular & 803.7 & 614.9 & 43.5 & 80.4 \\
 & & Integrated & 854.0 & 575.0 & 55.0 & 81.7 \\
\cmidrule{2-7}
 & \multirow{3}{*}{Jan\,11} & Point  & 774.6 & 545.2 & 45.7 & 79.2 \\
 & & Modular & 811.1 & 607.8 & 42.4 & 76.8 \\
 & & Integrated & 911.5 & 646.3 & 55.0 & 75.3 \\
\cmidrule{2-7}
 & \multirow{3}{*}{Jan\,12} & Point  & 992.4 & 680.6 & 45.8 & 85.6 \\
 & & Modular & 821.8 & 593.6 & 43.5 & 87.5 \\
 & & Integrated & 1004.8 & 734.2 & 55.2 & 84.5 \\
\midrule
\multirow{9}{*}{TFT}
 & \multirow{3}{*}{Jan\,10} & Point  & 122.9 & 108.9 & 10.7 & 96.5 \\
 & & Modular & 156.3 & 126.1 & 10.2 & 96.0 \\
 & & Integrated & 28.2 & 24.0 & 2.2 & 99.2 \\
\cmidrule{2-7}
 & \multirow{3}{*}{Jan\,11} & Point  & 184.5 & 163.9 & 13.0 & 93.7 \\
 & & Modular & 157.5 & 111.7 & 8.0 & 95.7 \\
 & & Integrated & 33.8 & 27.3 & 2.4 & 99.0 \\
\cmidrule{2-7}
 & \multirow{3}{*}{Jan\,12} & Point  & 373.6 & 270.2 & 19.5 & 94.3 \\
 & & Modular & 358.4 & 259.5 & 17.9 & 94.5 \\
 & & Integrated & 83.2 & 55.7 & 3.6 & 98.8 \\
\bottomrule
\end{tabular}
\end{table}

\begin{figure}[!htbp]
  \centering
  \includegraphics[width=.7\linewidth]{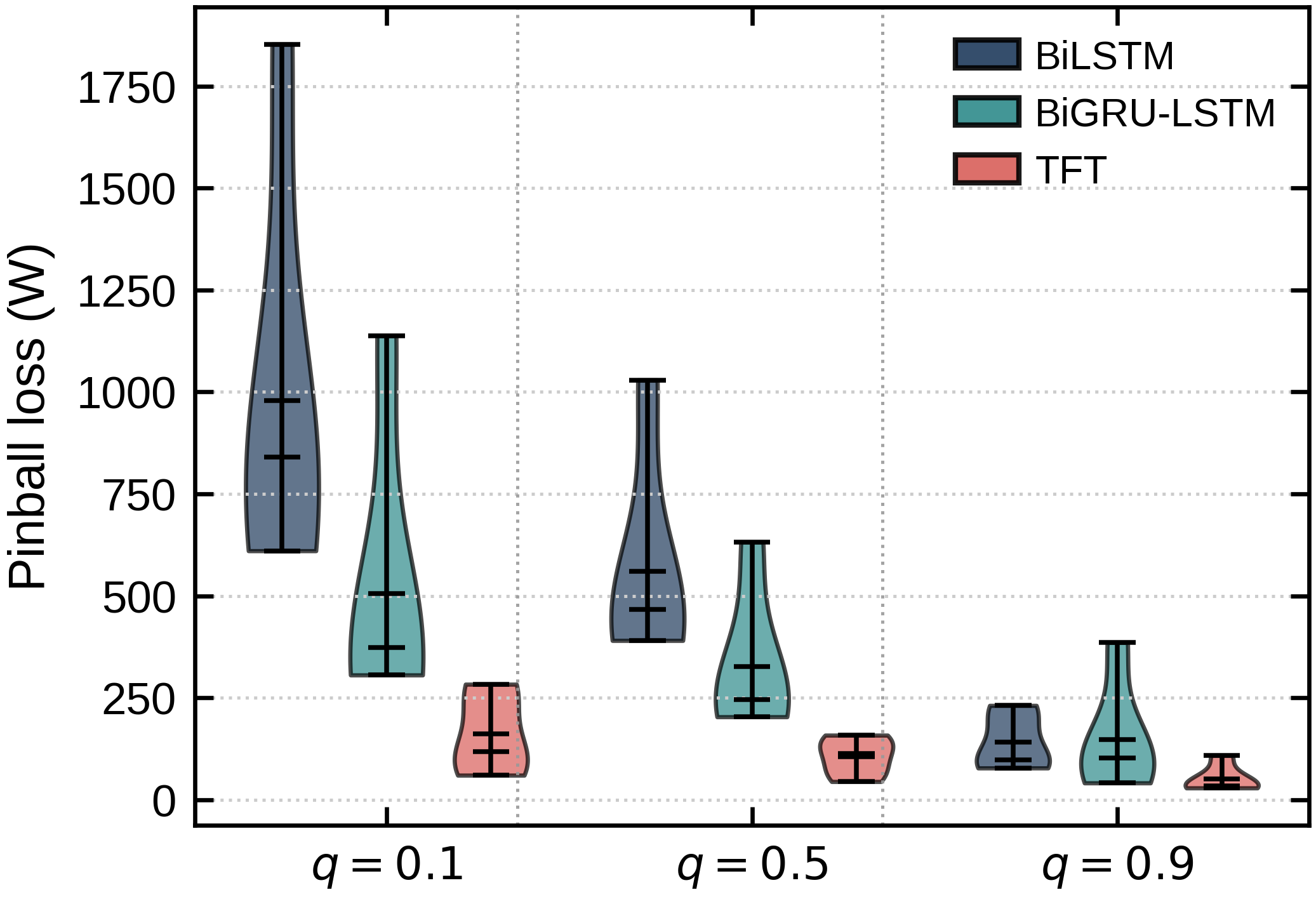}
  \caption{Pinball-loss distributions at $q\!\in\!\{0.1,0.5,0.9\}$ for Jan\,10-12, 2025. The TFT achieves the lowest and most stable losses.}
  \label{fig:residual_violin}
\end{figure}

\begin{figure}[!htbp]
\centering
\includegraphics[width=\columnwidth]{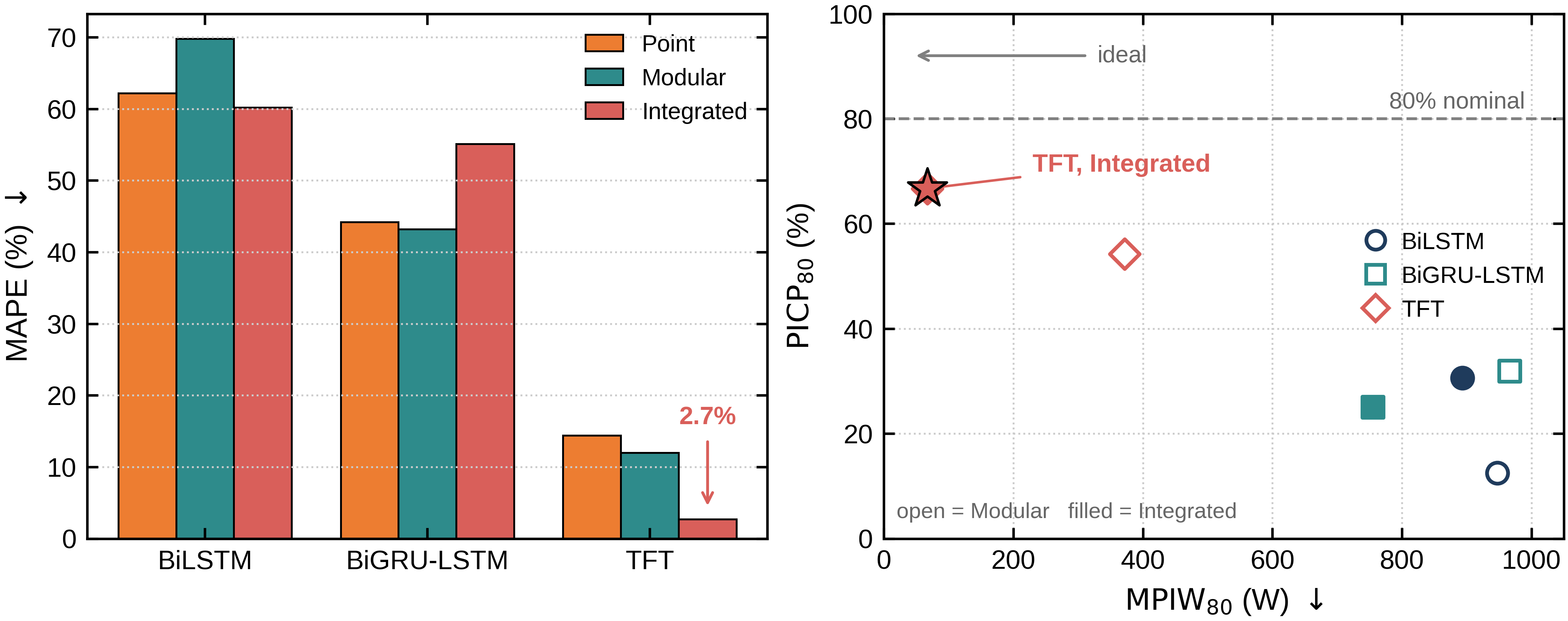}
\caption{\textcolor{black}{Mean point accuracy and interval trade-off across the three labeled days.}}
\label{fig:metrics_comparison}
\end{figure}

\paragraph{Interval Calibration and Sharpness}
\textcolor{black}{Operators need a confidence range, not only an expected value, so each model emits an 80\% interval that should contain the true load 80\% of the time; a good interval is calibrated with coverage near nominal and sharp, i.e., narrow at that coverage. Table~\ref{tab:prob_all} shows the integrated scheme yields more reliable intervals than the modular one, clearly for TFT. Under the modular scheme, the recurrent models produce wide, mis-centered intervals with coverage as low as 12.5\%. The integrated TFT is best with ${PICP}_{80}\approx67\%$ (75\% on individual days), ${MPIW}_{80}=67$\,W, and ${MIS}_{80}=176$\,W. Table~\ref{tab:perday_interval} confirms this on every labeled day.}
\begin{table}[!htbp]
\centering
\caption{Interval metrics for the modular and integrated schemes. \colorbox{goodcell}{Teal} marks the best per metric within each model. \textcolor{black}{Values are averages over the three labeled days (Jan\,10-12).}}
\label{tab:prob_all}
\begin{tabular}{llccc}
\toprule
Model & Method & ${PICP}_{80}\leftrightarrow$ (\%) & ${MPIW}_{80}\downarrow$ (W) & ${MIS}_{80}\downarrow$ (W) \\
\midrule
\multirow{2}{*}{BiLSTM} 
 & Modular    & 12.5 & 947.5 & 6661.0 \\
 & Integrated & \cellcolor{goodcell}30.6 & \cellcolor{goodcell}893.3 & \cellcolor{goodcell}4649.6 \\
\midrule
\multirow{2}{*}{BiGRU-LSTM} 
 & Modular    & \cellcolor{goodcell}31.9 & 966.6 & \cellcolor{goodcell}3422.4 \\
 & Integrated & 25.0 & \cellcolor{goodcell}755.5 & 4929.4 \\
\midrule
\multirow{2}{*}{TFT} 
 & Modular    & 54.2 & 372.0 & 859.2 \\
 & Integrated & \cellcolor{goodcell}66.7 & \cellcolor{goodcell}67.3 & \cellcolor{goodcell}175.7 \\
\bottomrule
\end{tabular}
\end{table}
\begin{table}[!htbp]
\centering
\caption{\textcolor{black}{Per-day interval metrics at the 80\% level on the labeled days.}}
\label{tab:perday_interval}
\begin{tabular}{lllccc}
\toprule
Model & Day & Method & ${PICP}_{80}\leftrightarrow$ & ${MPIW}_{80}\downarrow$ & ${MIS}_{80}\downarrow$ \\
\midrule
\multirow{6}{*}{BiLSTM}
 & \multirow{2}{*}{Jan\,10} & Modular & 4.2 & 970.9 & 7806.8 \\
 & & Integrated & 37.5 & 876.7 & 4000.6 \\
\cmidrule{2-6}
 & \multirow{2}{*}{Jan\,11} & Modular & 8.3 & 800.8 & 7917.6 \\
 & & Integrated & 29.2 & 866.3 & 4775.6 \\
\cmidrule{2-6}
 & \multirow{2}{*}{Jan\,12} & Modular & 25.0 & 1070.7 & 4258.6 \\
 & & Integrated & 25.0 & 937.0 & 5172.5 \\
\midrule
\multirow{6}{*}{BiGRU-LSTM}
 & \multirow{2}{*}{Jan\,10} & Modular & 29.2 & 985.7 & 3906.6 \\
 & & Integrated & 29.2 & 765.0 & 4302.7 \\
\cmidrule{2-6}
 & \multirow{2}{*}{Jan\,11} & Modular & 33.3 & 887.6 & 3560.8 \\
 & & Integrated & 25.0 & 716.9 & 4988.1 \\
\cmidrule{2-6}
 & \multirow{2}{*}{Jan\,12} & Modular & 33.3 & 1026.5 & 2799.8 \\
 & & Integrated & 20.8 & 784.5 & 5497.4 \\
\midrule
\multirow{6}{*}{TFT}
 & \multirow{2}{*}{Jan\,10} & Modular & 54.2 & 396.4 & 676.0 \\
 & & Integrated & 62.5 & 48.0 & 117.5 \\
\cmidrule{2-6}
 & \multirow{2}{*}{Jan\,11} & Modular & 70.8 & 357.6 & 647.5 \\
 & & Integrated & 75.0 & 50.2 & 123.7 \\
\cmidrule{2-6}
 & \multirow{2}{*}{Jan\,12} & Modular & 37.5 & 362.1 & 1254.2 \\
 & & Integrated & 62.5 & 103.6 & 285.8 \\
\bottomrule
\end{tabular}
\end{table}
\textcolor{black}{The evaluation uses a single 80\% level because it matches the emitted quantiles $(q_{0.1},q_{0.5},q_{0.9})$; other levels would require retraining with matching quantiles, which we leave to future work. The level choice does not bias the ranking: ${MIS}_{80}$ already weighs calibration against sharpness by construction, and the distribution-level QS agrees, again ranking the integrated TFT first at $11.8$\,W versus 273-278\,W for the recurrent models (Fig.~\ref{fig:residual_violin}).}
\textcolor{black}{Fig.~\ref{fig:metrics_comparison} combines both views: (a) point accuracy by method and (b) the coverage-width trade-off, where the ideal sits at 80\% coverage with small width. The integrated TFT lies closest to this ideal; the recurrent models under-cover despite wider intervals.}

\paragraph{Statistical Significance}
\textcolor{black}{The rankings above rest on metric differences over only 72 labeled hours (three days $\times$ 24\,h), so we test the schemes with the Diebold-Mariano (DM) test under the Harvey-Leybourne-Newbold (HLN) small-sample correction. Table~\ref{tab:dm} shows that a negative statistic favors the integrated scheme. Comparing the schemes on pinball losses, integrated training significantly improves TFT (${DM}=-7.00$, $p<10^{-8}$) and BiLSTM (${DM}=-2.70$, $p=0.009$), while BiGRU-LSTM favors the modular pipeline (${DM}=+2.20$, $p=0.031$), consistent with its mixed interval behavior. Comparing the models under the integrated scheme on squared-error losses, TFT significantly outperforms BiLSTM (${DM}=-4.50$, $p\approx2\times10^{-5}$) and BiGRU-LSTM (${DM}=-4.50$, $p\approx3\times10^{-5}$), and BiGRU-LSTM outperforms BiLSTM (${DM}=-3.90$, $p\approx2\times10^{-4}$).}
\begin{table}[!htbp]
\centering
\caption{\textcolor{black}{DM tests over the labeled period ($n=72$ hours, HLN-corrected). Top: integrated vs.\ modular using pinball losses, where negative values favor the integrated scheme. Bottom: between-model comparisons under the integrated scheme using squared-error losses, where negative values favor the first-named model.}}
\label{tab:dm}
\begin{tabular}{lcc}
\toprule
Comparison & DM statistic & $p$-value \\
\midrule
\multicolumn{3}{c}{\textit{Integrated vs.\ modular}} \\
\midrule
BiLSTM      & $-2.70$ & $0.009$ \\
BiGRU-LSTM  & $+2.20$ & $0.031$ \\
TFT         & $-7.00$ & $<10^{-8}$ \\
\midrule
\multicolumn{3}{c}{\textit{Models under integrated scheme}} \\
\midrule
TFT vs.\ BiLSTM        & $-4.50$ & $\approx2\times10^{-5}$ \\
TFT vs.\ BiGRU-LSTM    & $-4.50$ & $\approx3\times10^{-5}$ \\
BiGRU-LSTM vs.\ BiLSTM & $-3.90$ & $\approx2\times10^{-4}$ \\
\bottomrule
\end{tabular}
\end{table}

\subsubsection{Operational Behavior}
\begin{figure*}[!htbp]
\centering
\begin{subfigure}[t]{0.48\textwidth}\centering
    \includegraphics[width=\linewidth]{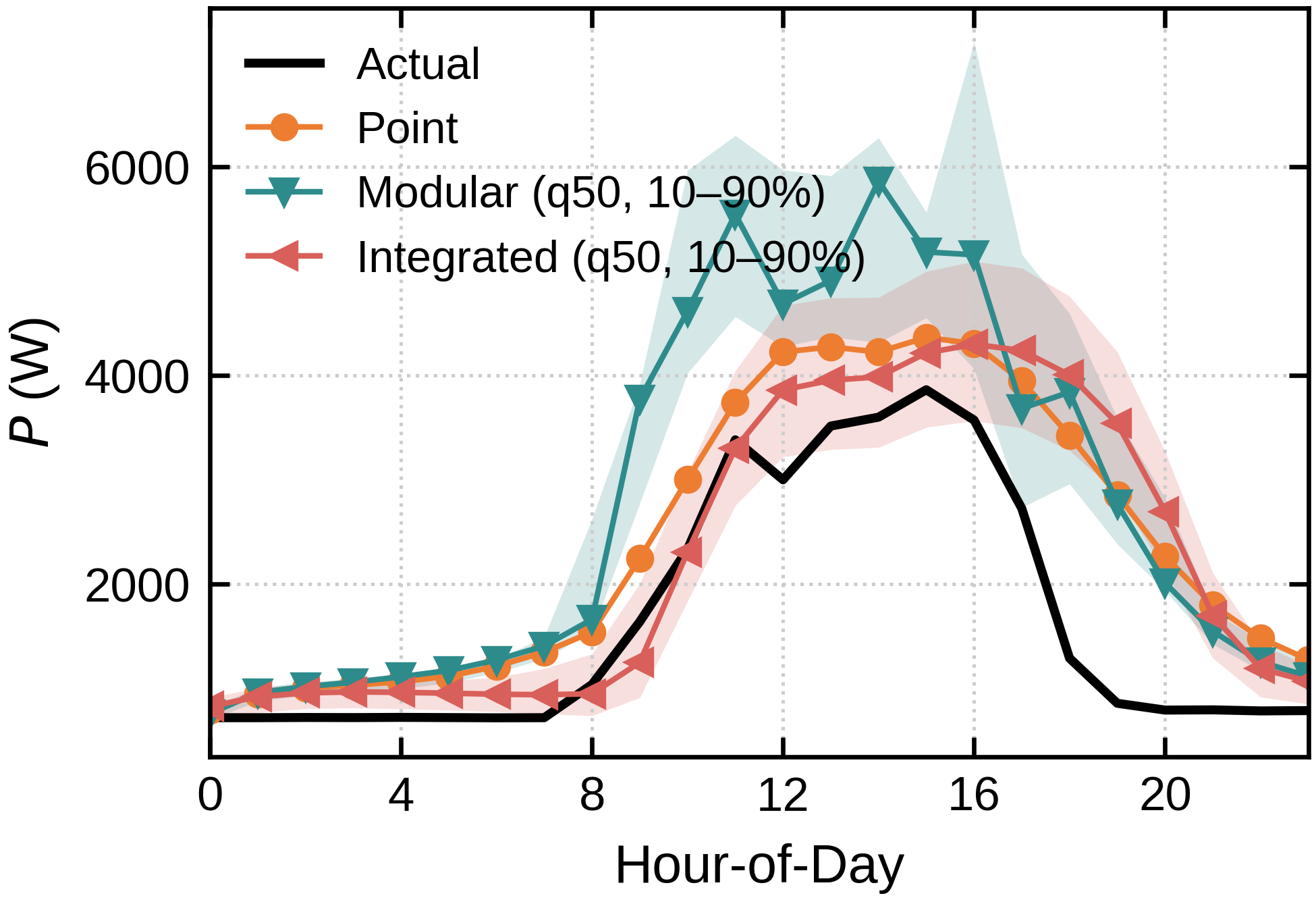}\caption{BiLSTM Jan\,10}
\end{subfigure}\hspace{.1em}
\begin{subfigure}[t]{0.48\textwidth}\centering
    \includegraphics[width=\linewidth]{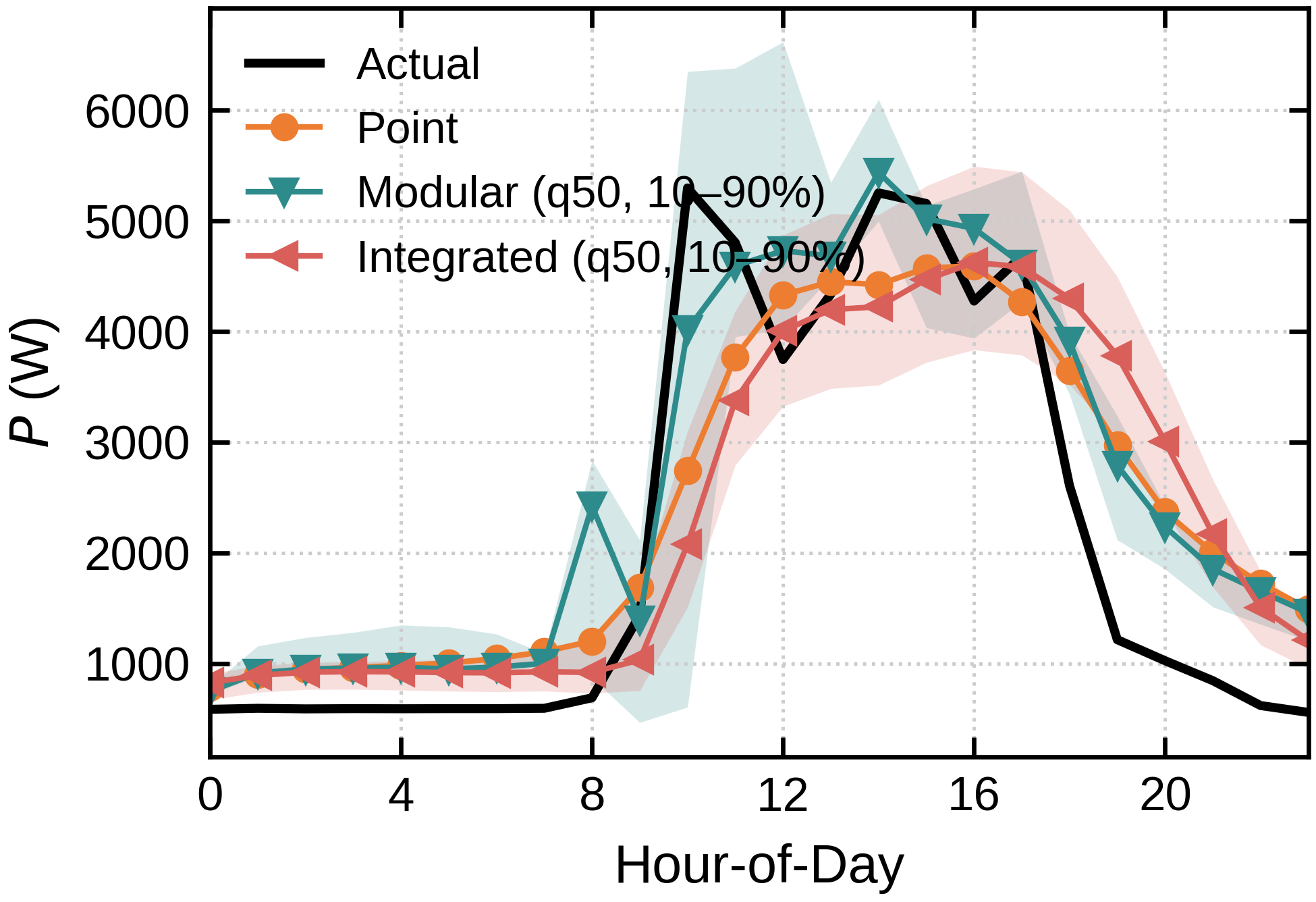}\caption{BiLSTM Jan\,12}
\end{subfigure}\vspace{-.5em}
  
\begin{subfigure}[t]{0.48\textwidth}\centering
    \includegraphics[width=\linewidth]{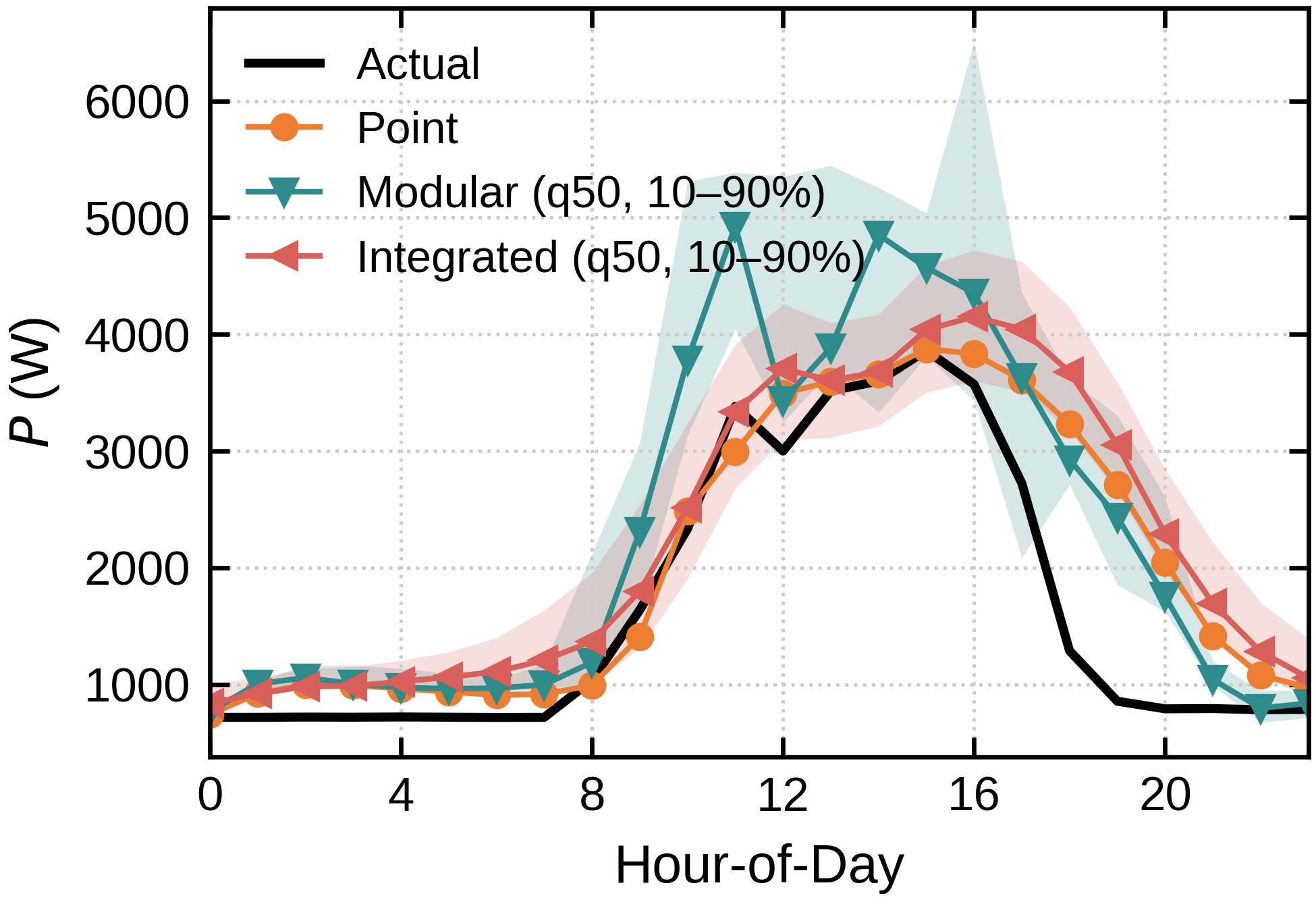}\caption{BiGRU-LSTM Jan\,10}
\end{subfigure}\hspace{.1em}
\begin{subfigure}[t]{0.48\textwidth}\centering
    \includegraphics[width=\linewidth]{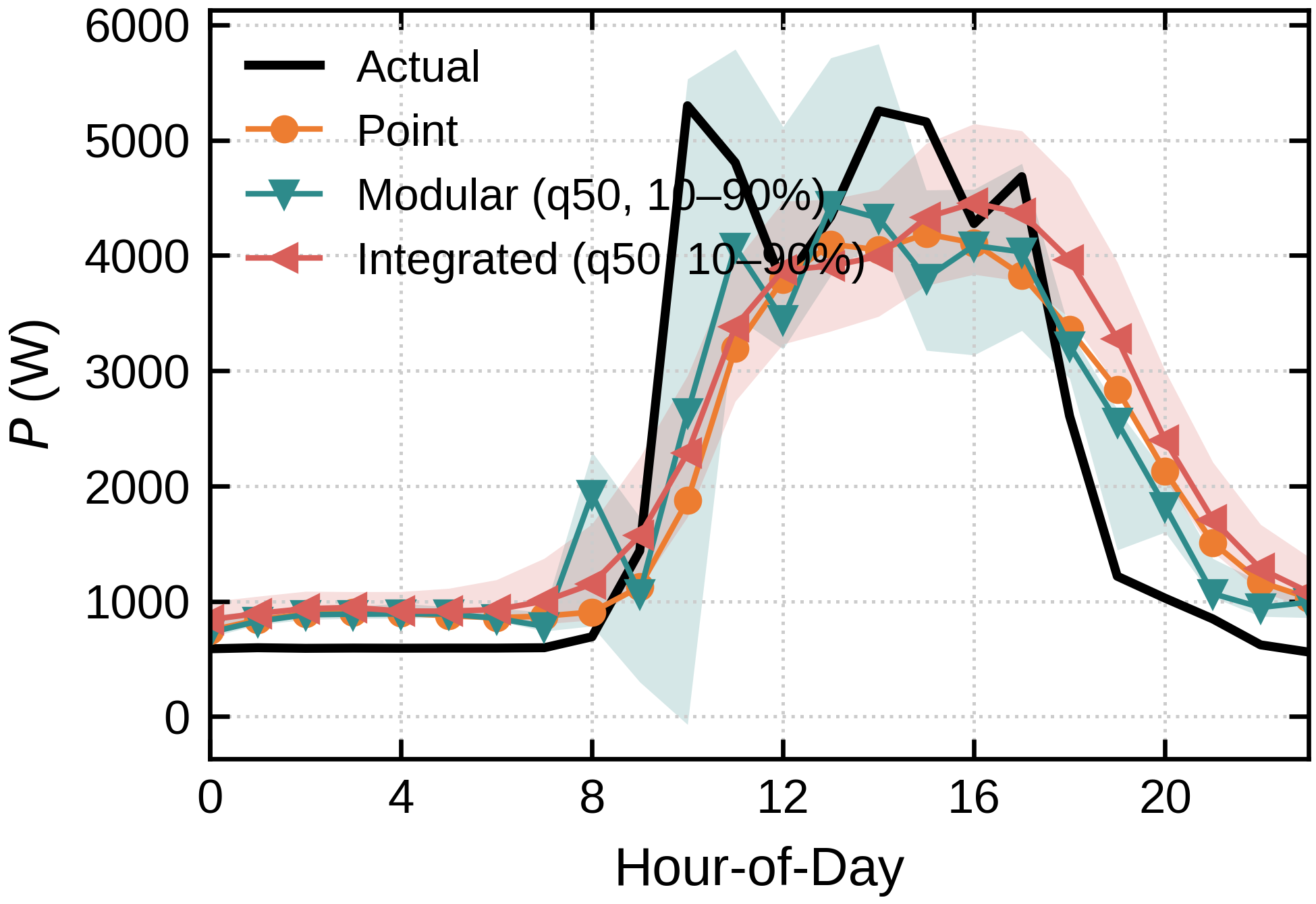}\caption{BiGRU-LSTM Jan\,12}
\end{subfigure}\vspace{-.5em}

\begin{subfigure}[t]{0.48\textwidth}\centering
    \includegraphics[width=\linewidth]{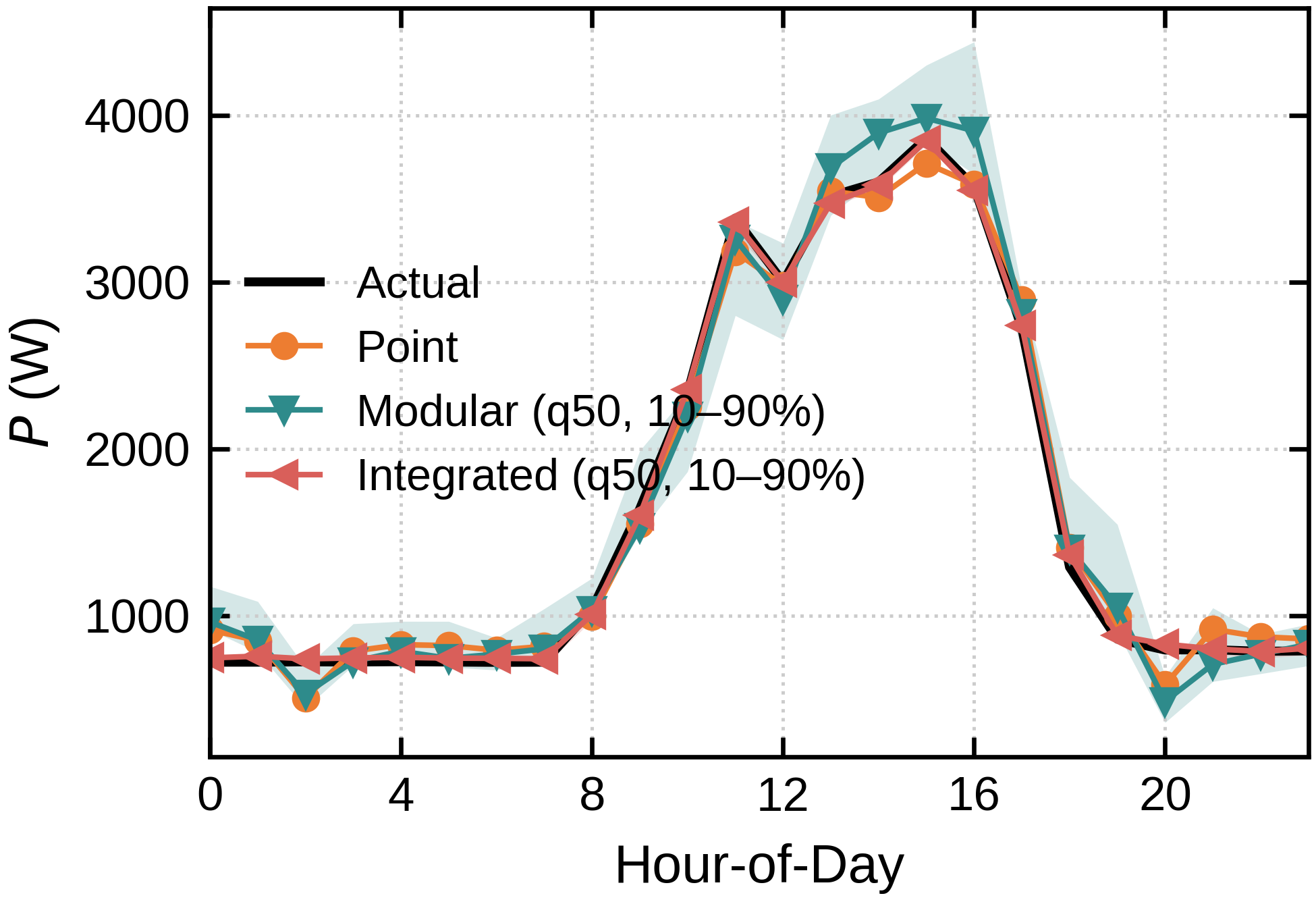}\caption{TFT Jan\,10}
\end{subfigure}\hspace{.1em}
\begin{subfigure}[t]{0.48\textwidth}\centering
    \includegraphics[width=\linewidth]{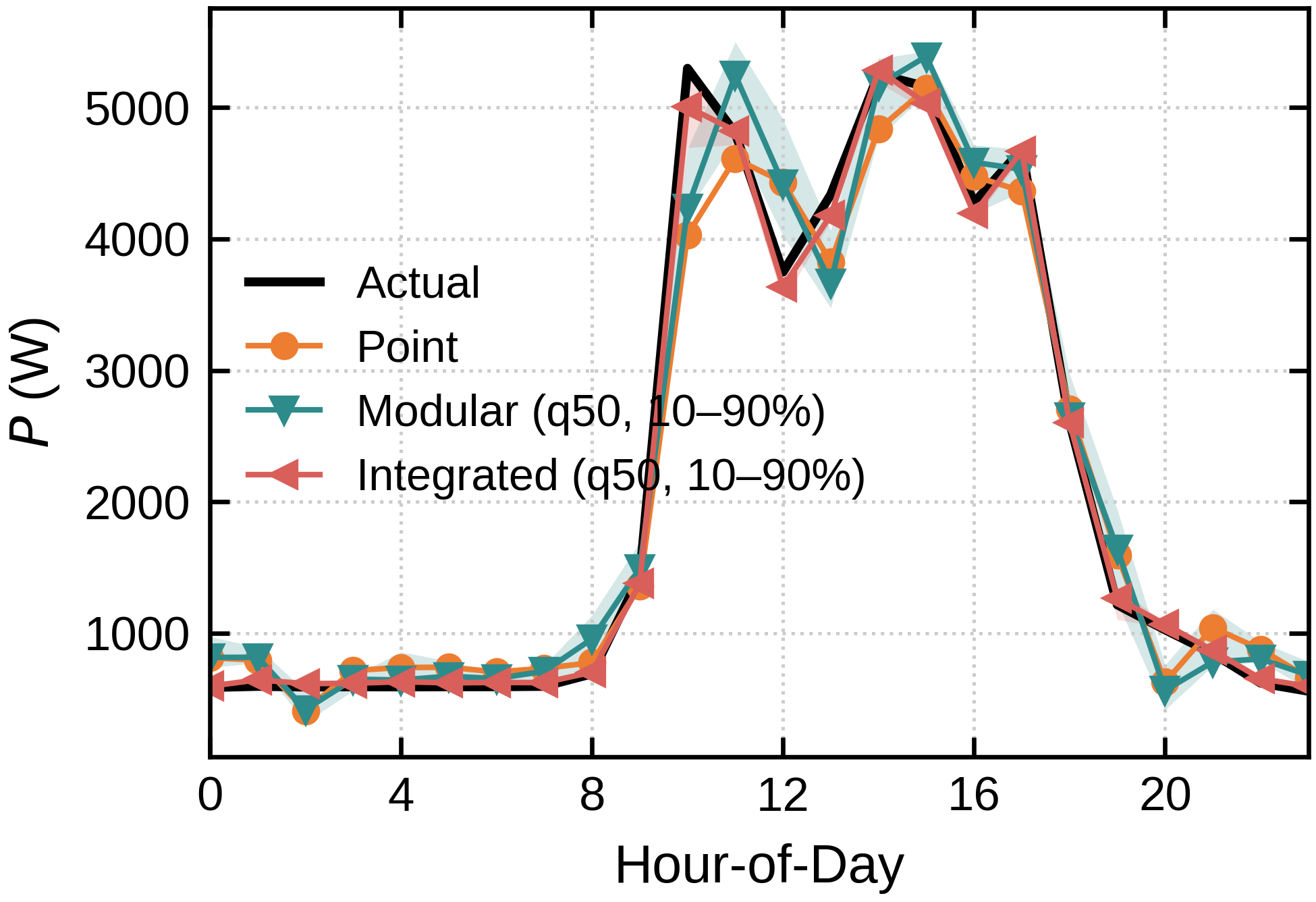}\caption{TFT Jan\,12}
\end{subfigure}\vspace{-.5em}
\caption{Actual vs.\ predicted demand on a moderate day (Jan\,10) and a high-stress day (Jan\,12), 2025. Each sub-figure shows the median forecast and the 10-90\% interval for one model.}\label{fig:daily-grid-15}
\end{figure*}

\begin{figure}[!htbp]
  \centering
  \includegraphics[width=.9\columnwidth]{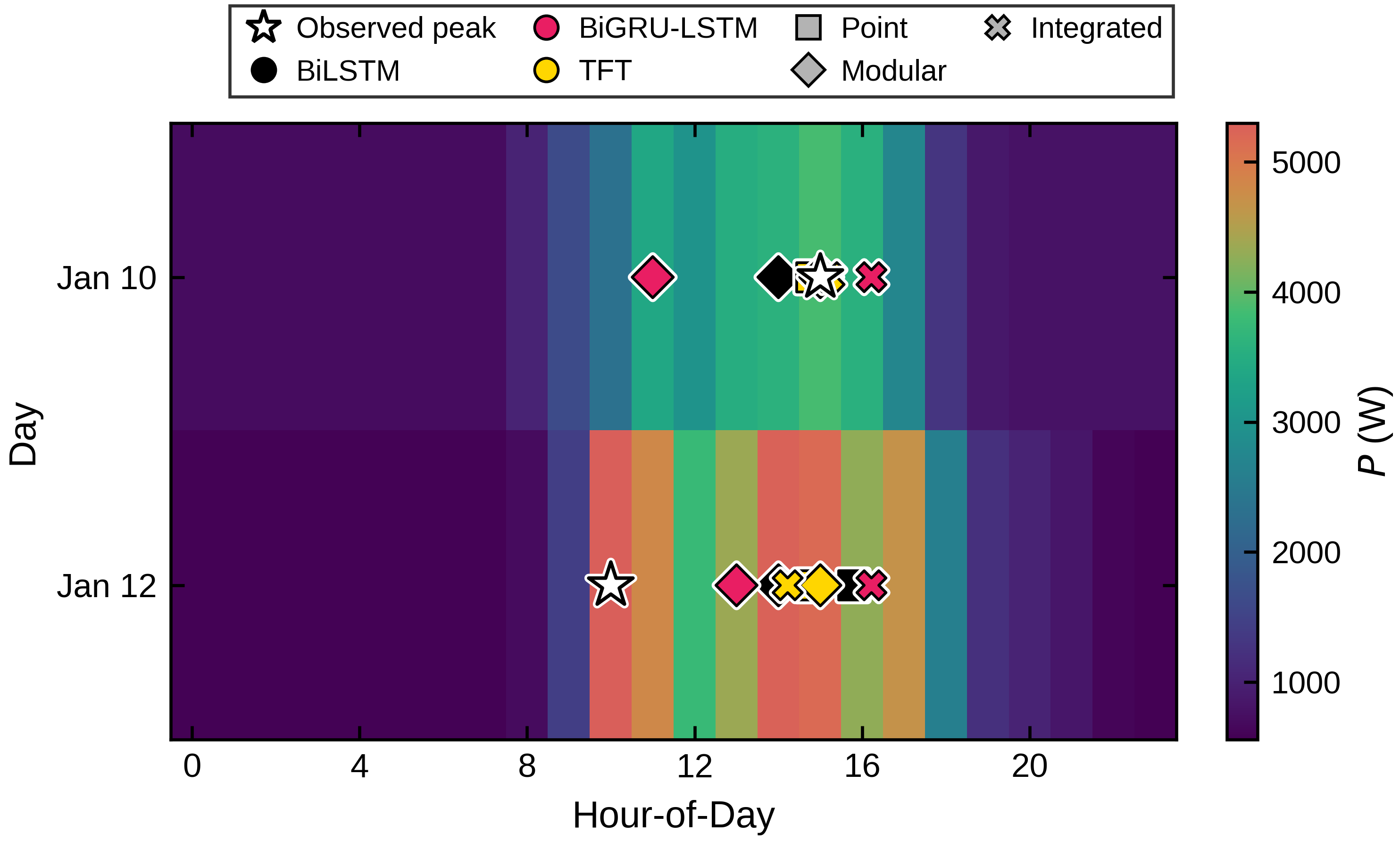}
  \caption{Actual demand over the $hod \times dow$ grid, with each model's predicted peak marked. The closer a marker sits to the actual peak cell, the better the peak timing.}
  \label{fig:hovmoller_peaks}
\end{figure}

\begin{figure*}[!htbp]
  \centering
  \begin{subfigure}[t]{.3\textwidth}\centering
    \includegraphics[width=\linewidth]{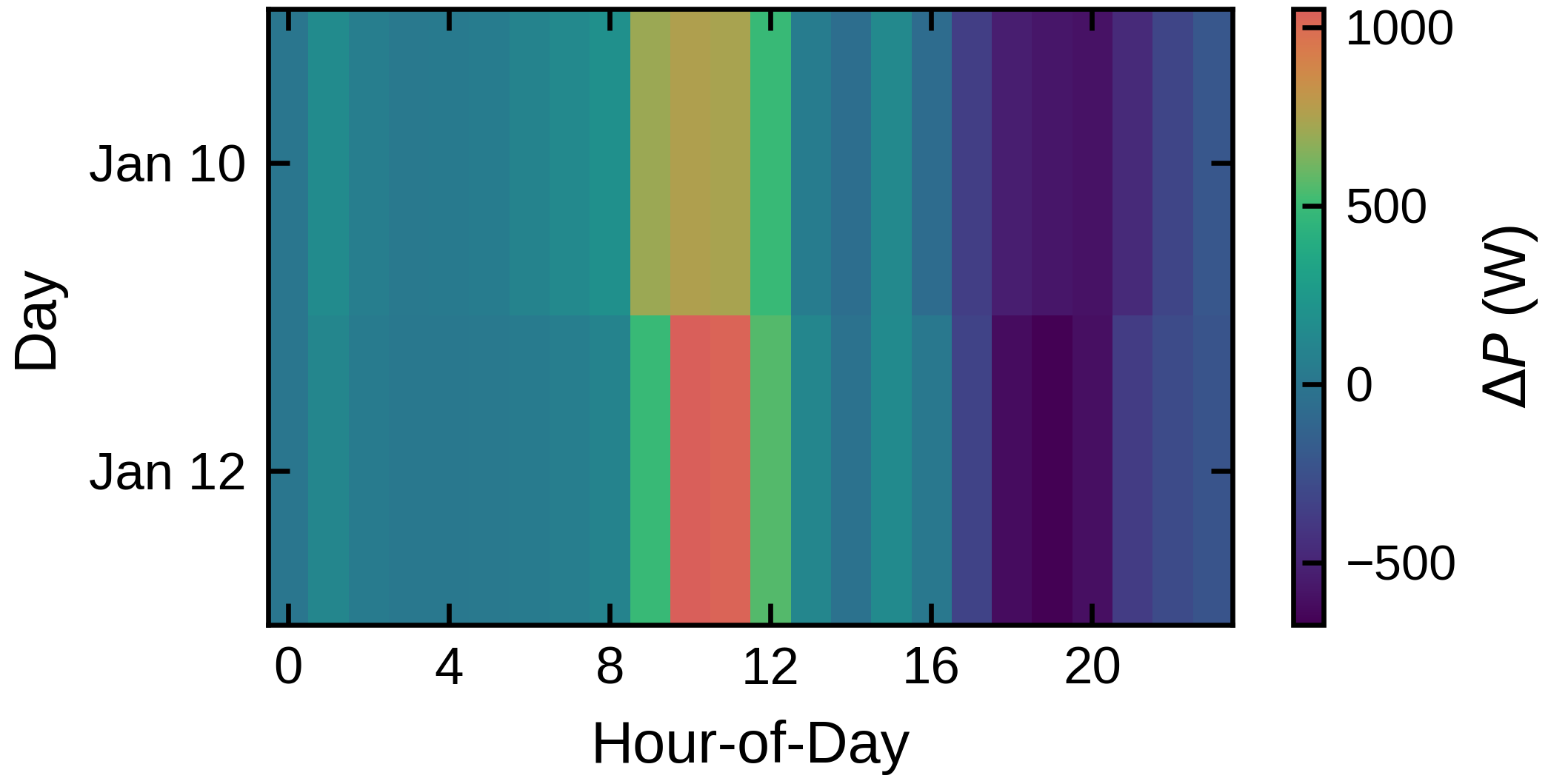}\caption{BiLSTM\,(Point)}
  \end{subfigure}\hspace{.1em}
  \begin{subfigure}[t]{0.3\textwidth}\centering
    \includegraphics[width=\linewidth]{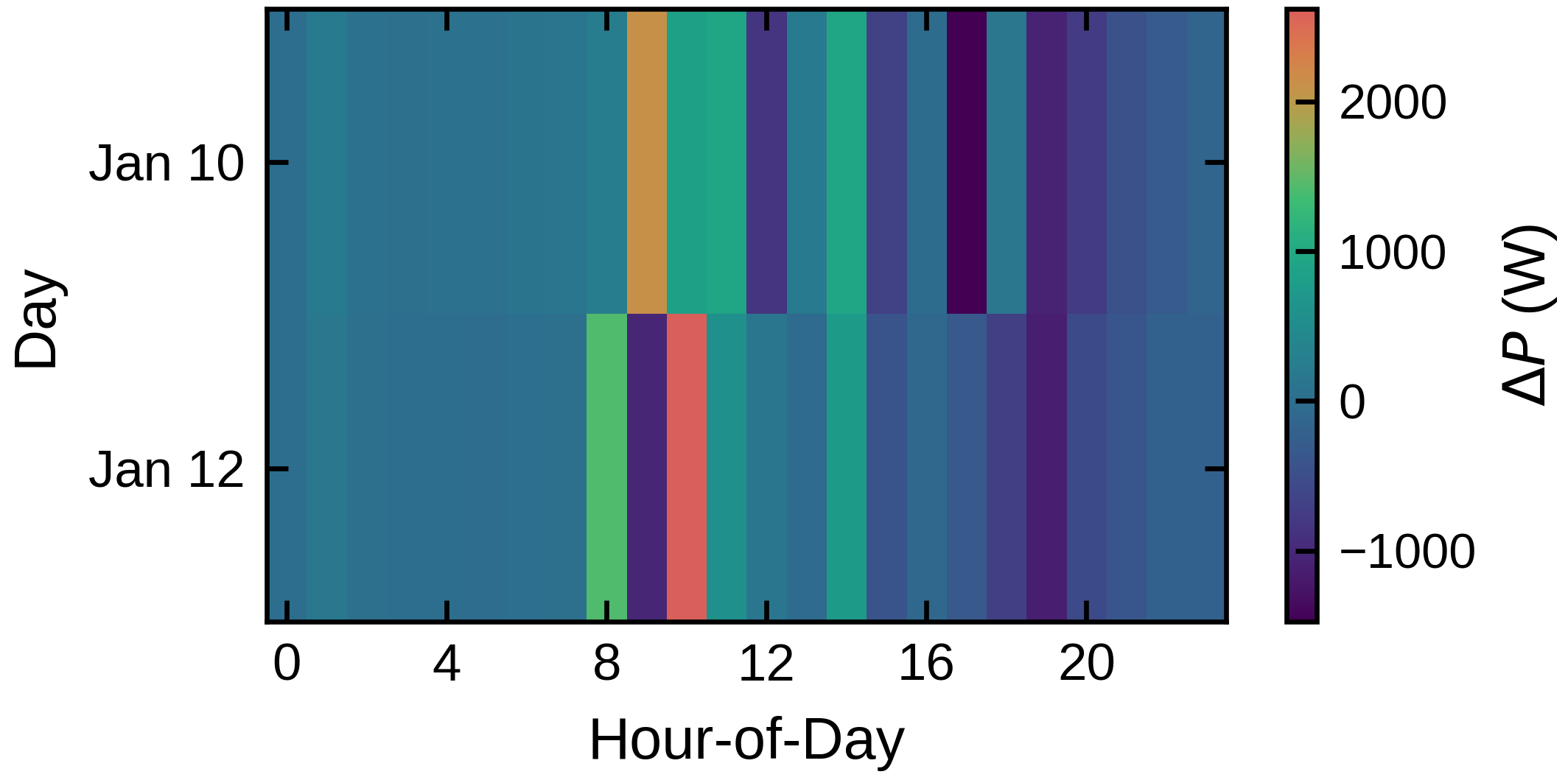}\caption{BiLSTM\,(Modular)}
  \end{subfigure}\hspace{.1em}
  \begin{subfigure}[t]{0.3\textwidth}\centering
    \includegraphics[width=\linewidth]{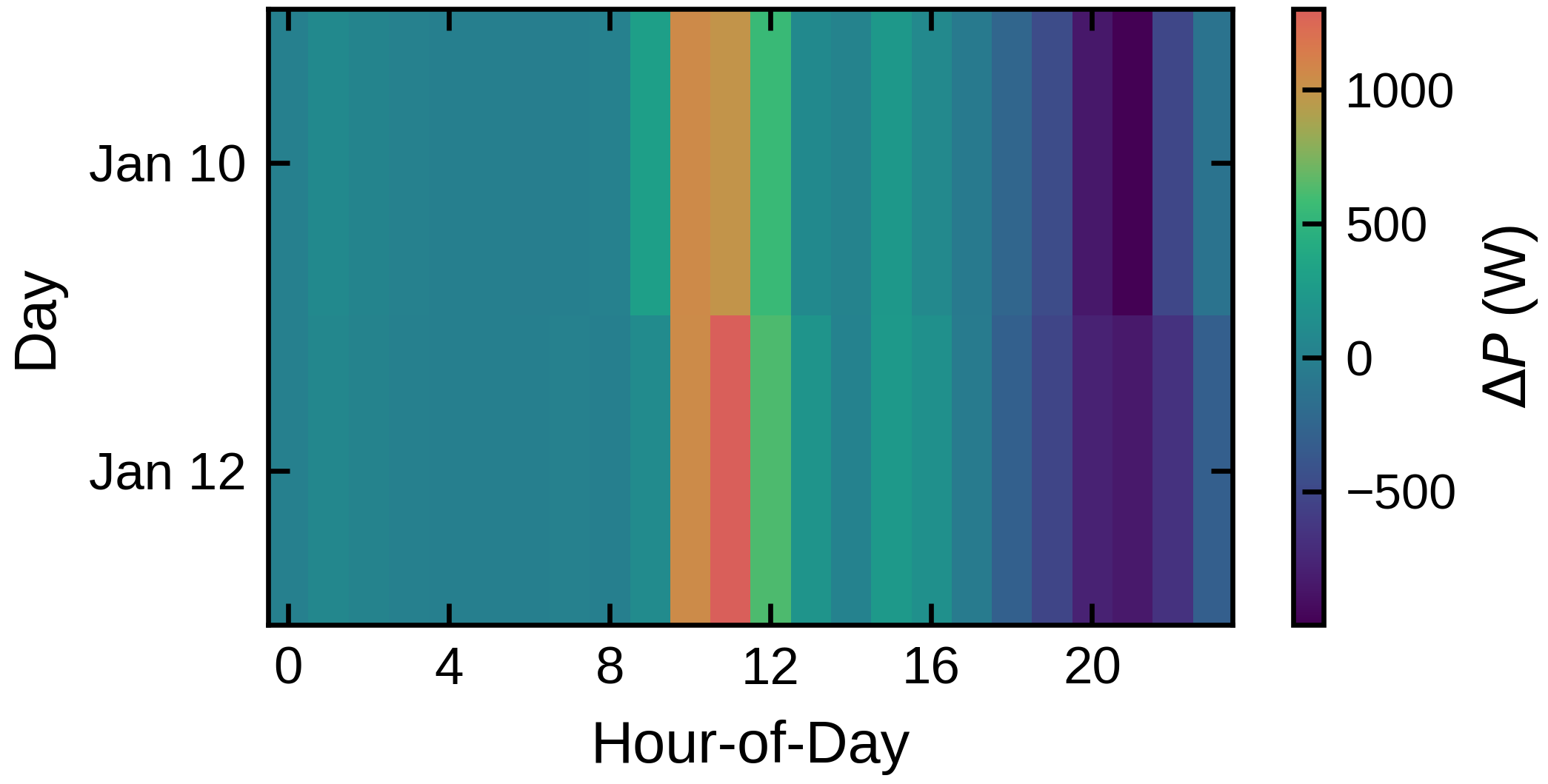}\caption{BiLSTM\,(Integrated)}
  \end{subfigure}\vspace{-.5em}
  \begin{subfigure}[t]{0.3\textwidth}\centering
    \includegraphics[width=\linewidth]{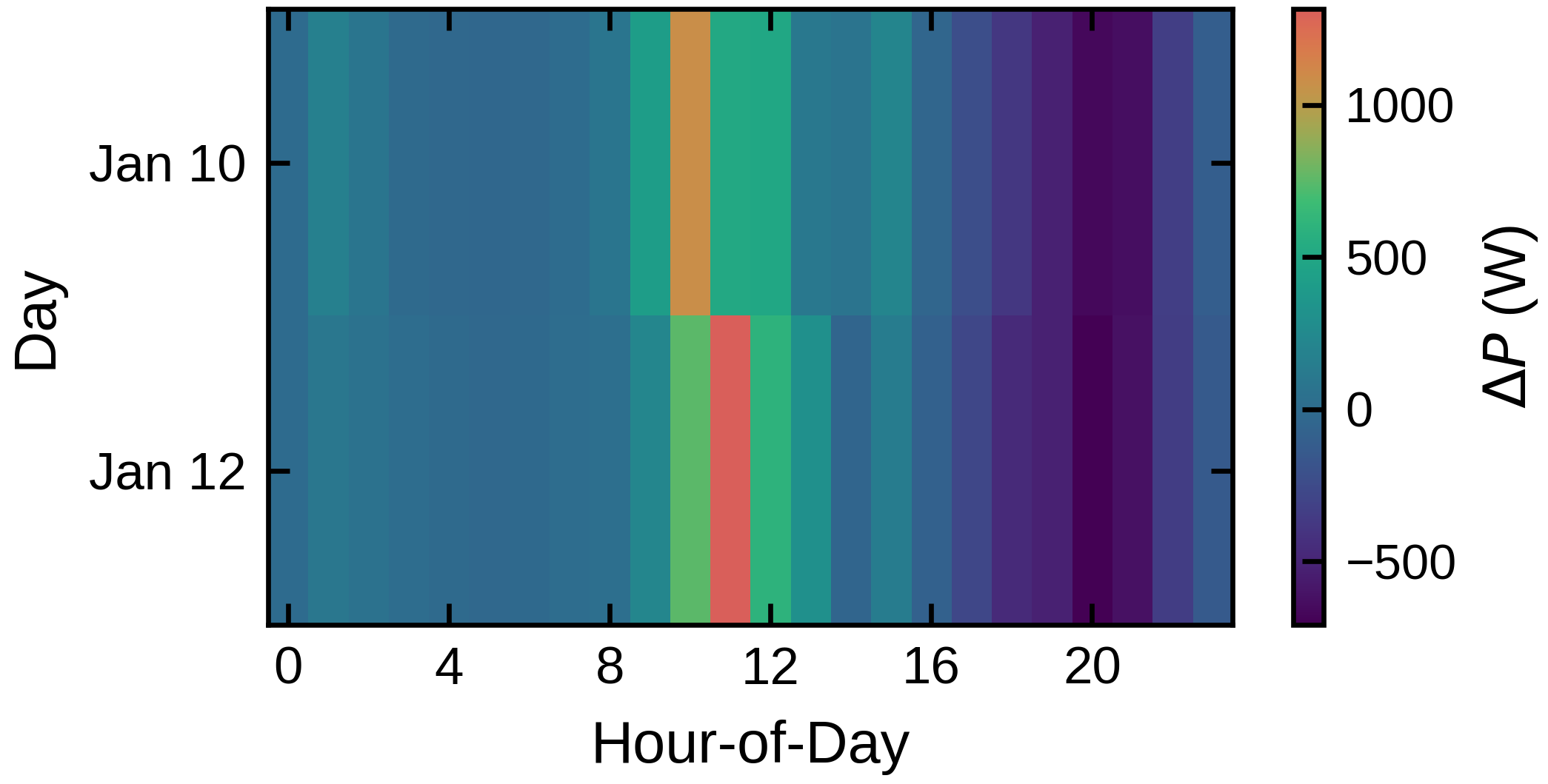}\caption{BiGRU-LSTM\,(Point)}
  \end{subfigure}\hspace{.1em}
  \begin{subfigure}[t]{0.3\textwidth}\centering
    \includegraphics[width=\linewidth]{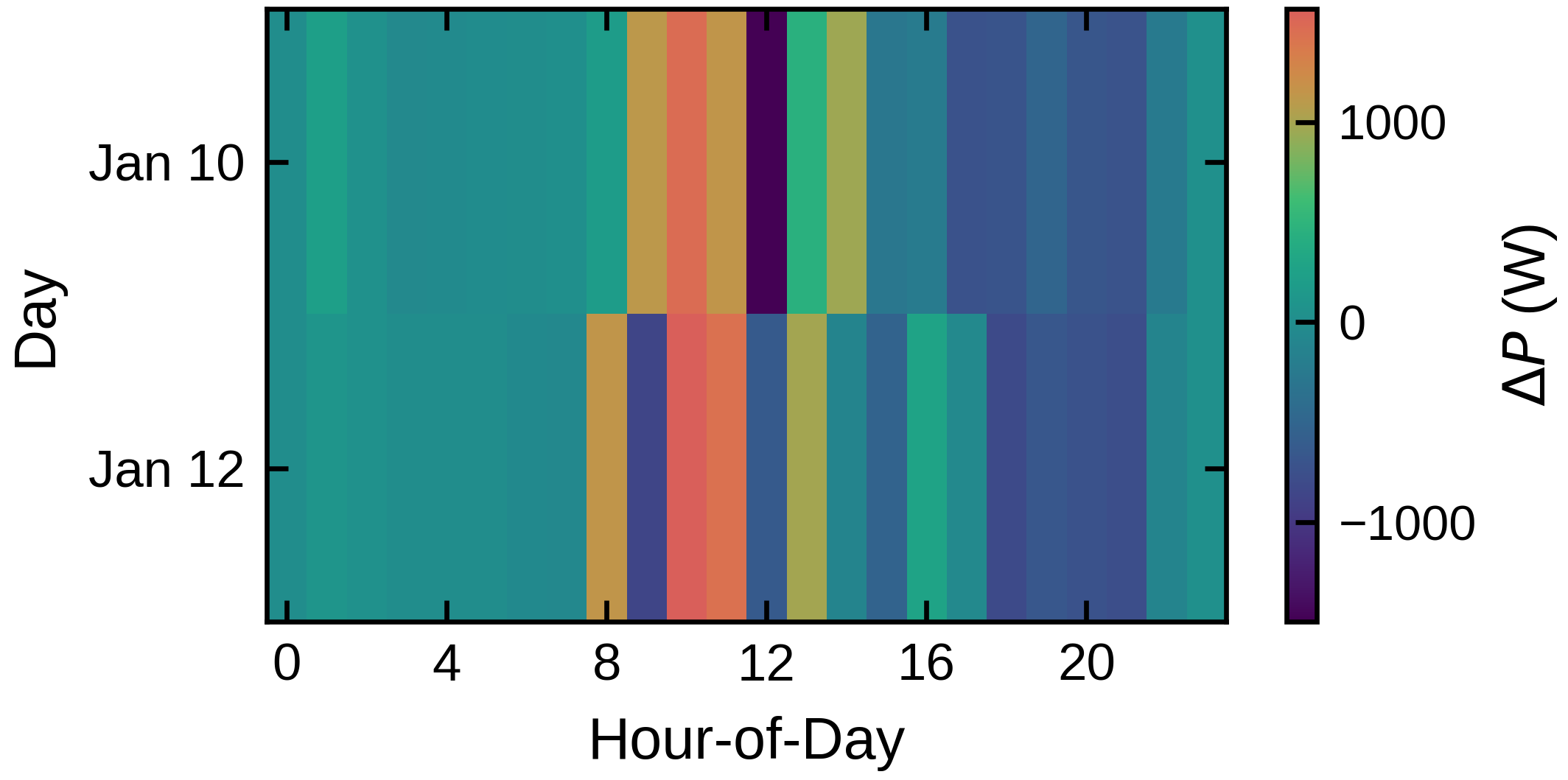}\caption{BiGRU-LSTM\,(Modular)}
  \end{subfigure}\hspace{.1em}
  \begin{subfigure}[t]{0.3\textwidth}\centering
    \includegraphics[width=\linewidth]{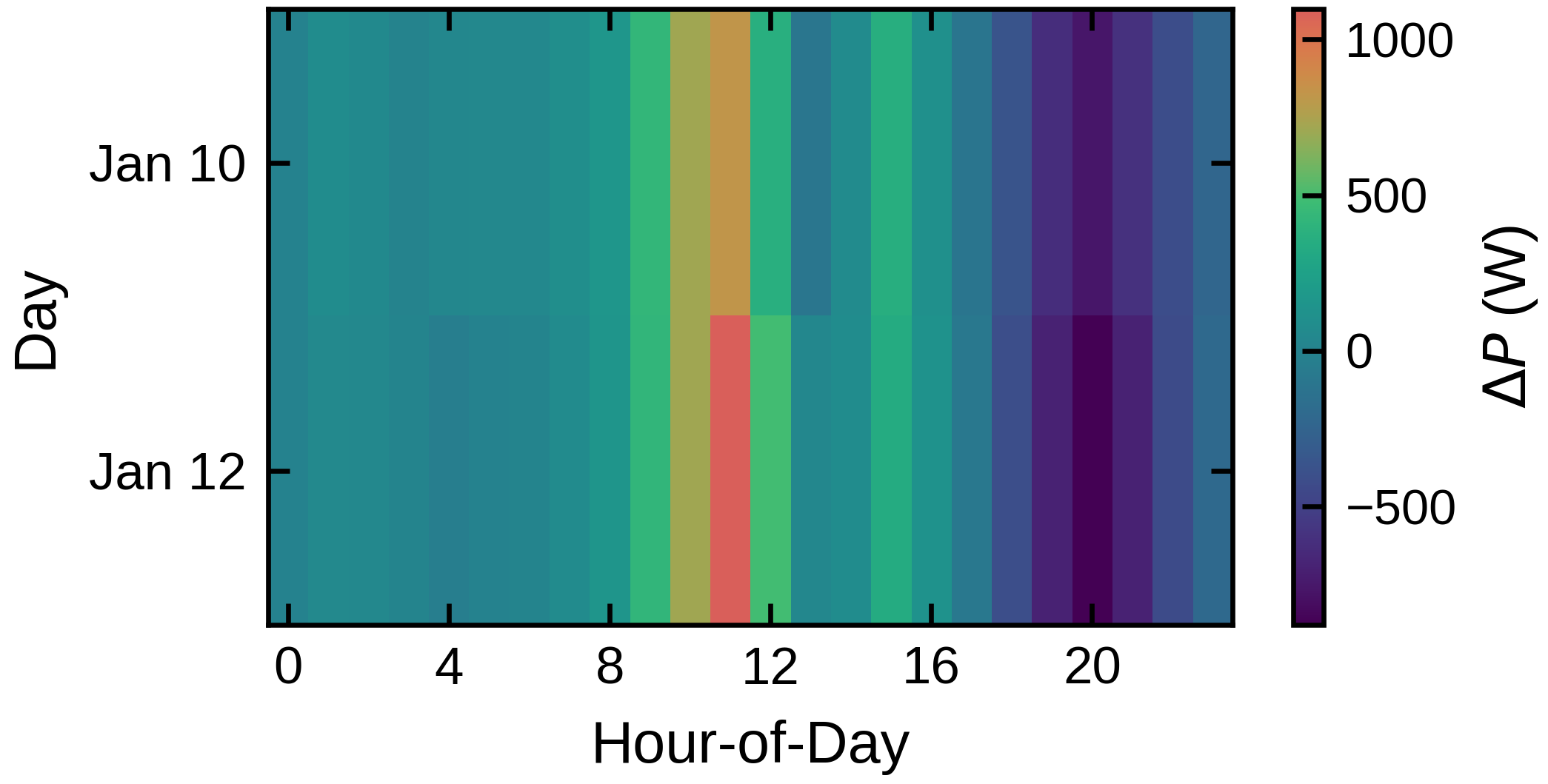}\caption{BiGRU-LSTM\,(Integrated)}
  \end{subfigure}\vspace{-.5em}
  \begin{subfigure}[t]{0.3\textwidth}\centering
    \includegraphics[width=\linewidth]{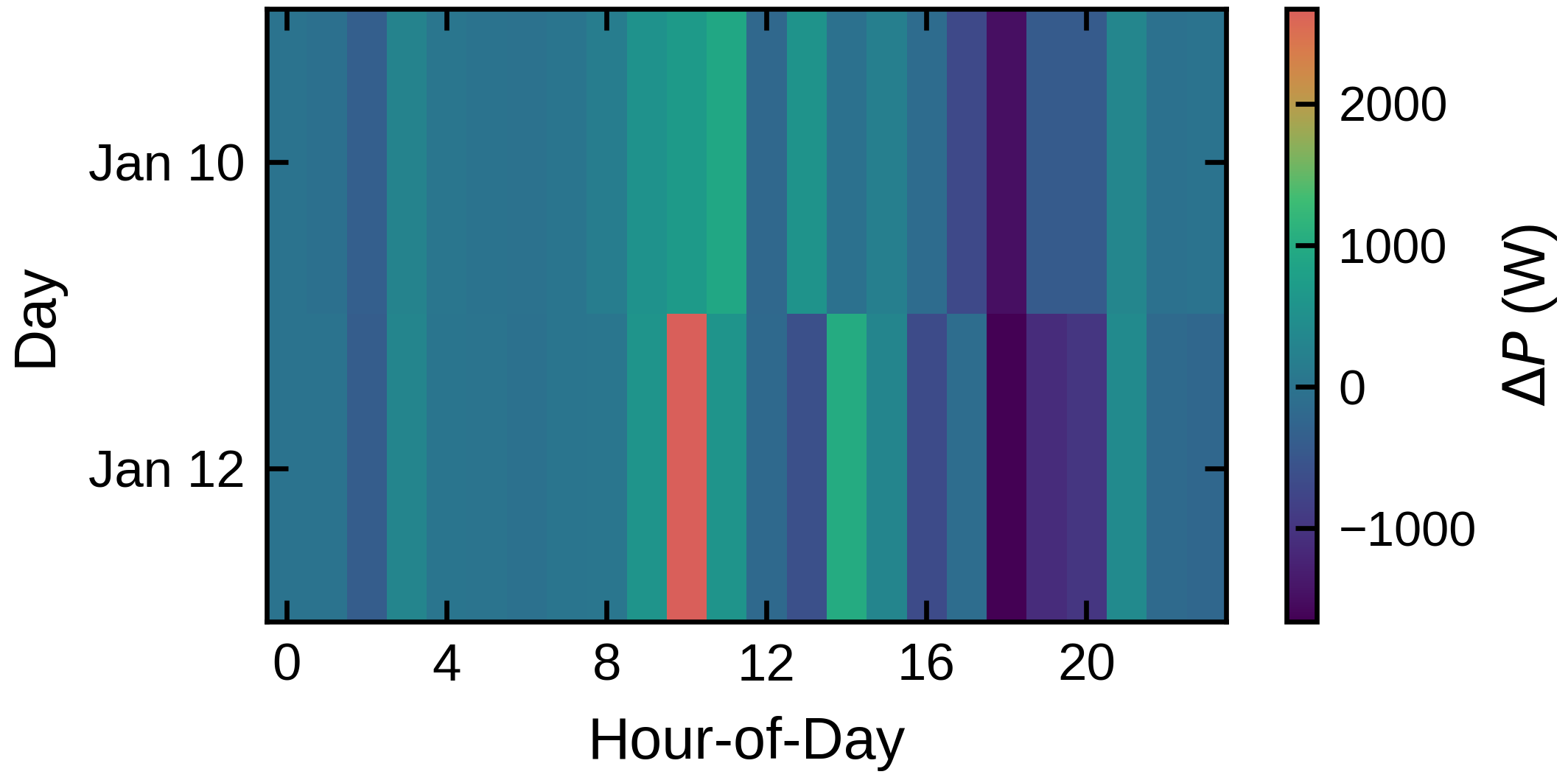}\caption{TFT\,(Point)}
  \end{subfigure}\hspace{.1em}
  \begin{subfigure}[t]{0.3\textwidth}\centering
    \includegraphics[width=\linewidth]{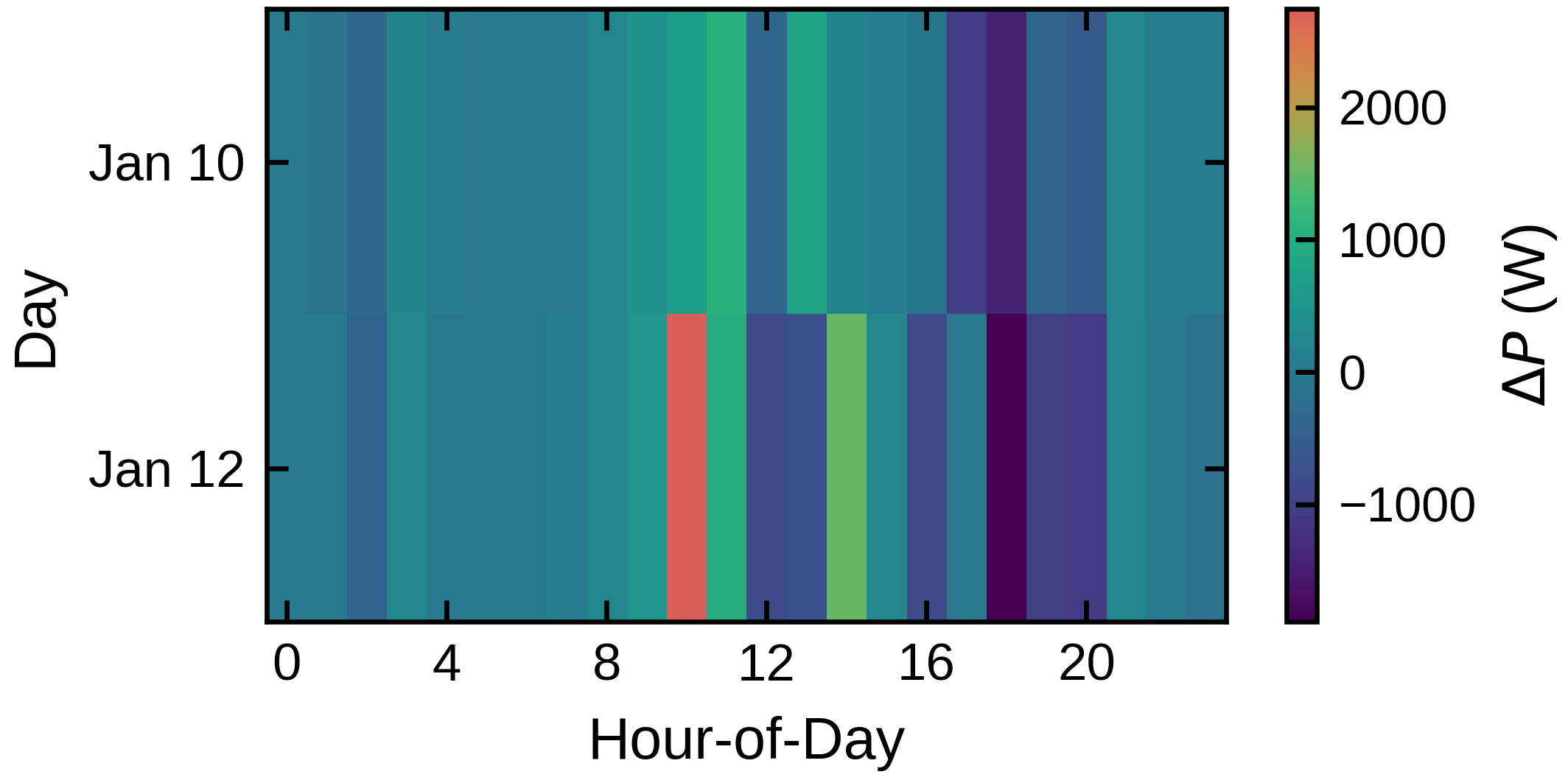}\caption{TFT\,(Modular)}
  \end{subfigure}\hspace{.1em}
  \begin{subfigure}[t]{0.3\textwidth}\centering
    \includegraphics[width=\linewidth]{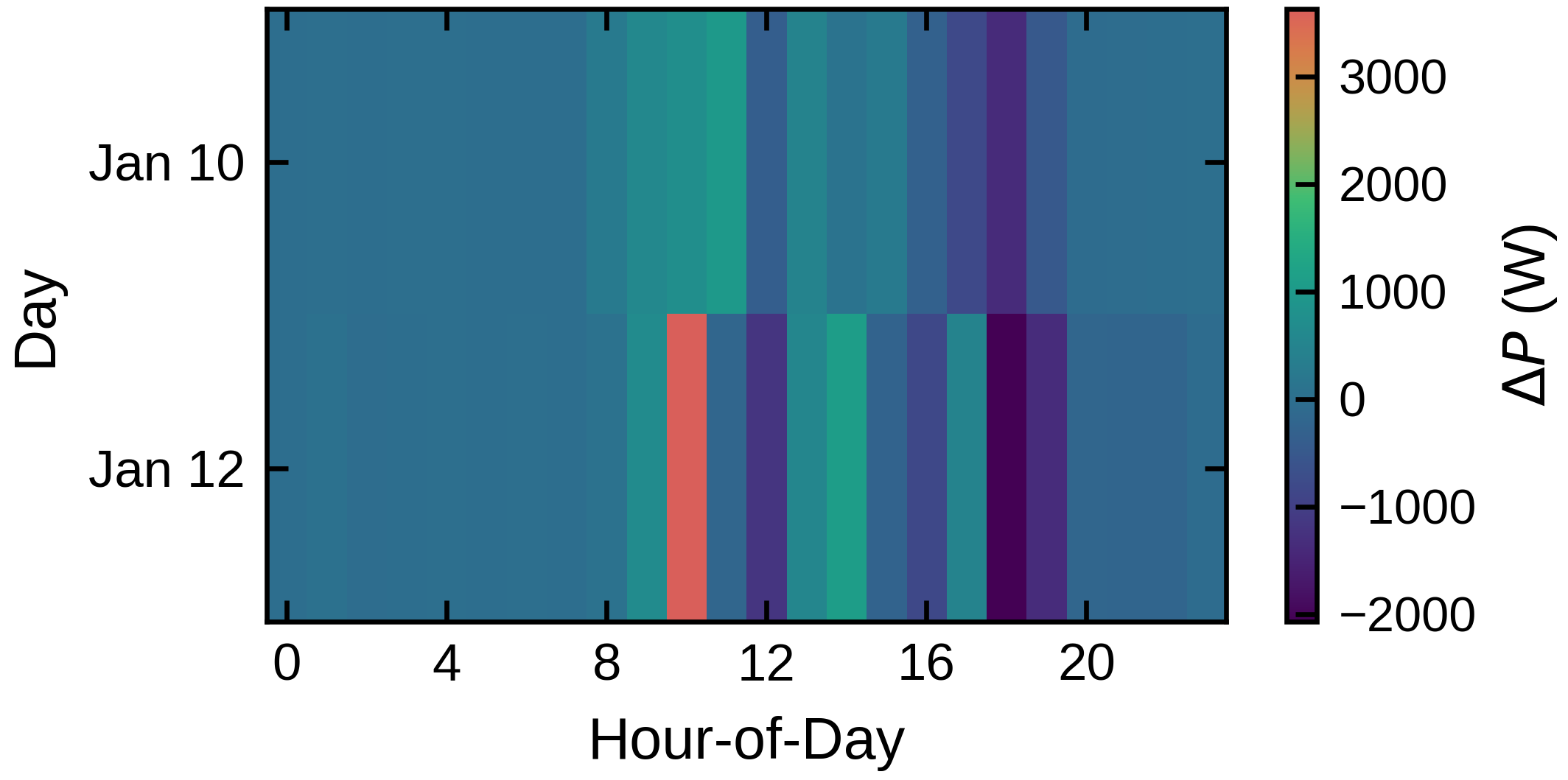}\caption{TFT\,(Integrated)}
  \end{subfigure}
  \caption{Hour-to-hour ramp rate ($\Delta P$) for the actual demand and each model-method combination. Diverging colors show up- vs.\ down-ramps.}
  \label{fig:ramp_rate}
\end{figure*}

\paragraph{Daily Pattern, Peak, and Ramp Reproduction}
\textcolor{black}{We next check how each model reproduces the demand curve. Fig.~\ref{fig:daily-grid-15} plots the median forecast and 80\% interval against the actual load on a moderate day (Jan\,10) and a high-stress day (Jan\,12). On the moderate day all three models track the daily shape, overshooting the peak by different margins, about 11\% for BiLSTM, 8\% for BiGRU-LSTM, under 1\% for TFT. TFT's interval adapts, narrow over the calm overnight hours (about 50\,W) and widening to several hundred Watts around the stress-day peak.}
\textcolor{black}{Zooming in on the peak, Table~\ref{tab:peak_deltas_grouped} gives the peak-timing ($\Delta h$) and magnitude ($\Delta P$) errors; Fig.~\ref{fig:hovmoller_peaks} and Fig.~\ref{fig:ramp_rate} show where peaks fall and how ramps compare. On the moderate day all models find the peak hour, with small magnitude errors, BiGRU-LSTM and TFT within about 150\,W, BiLSTM overshooting by $\approx$500\,W. The stress day separates them more clearly; the recurrent models miss the peak by 4-6\,h and underestimate it by over 1\,kW, falling $-61\%$ and $-57\%$ short at the true peak hour, while TFT holds timing to about 1\,h and magnitude to a few percent. The ramp maps in Fig.~\ref{fig:ramp_rate} show that TFT reproduces the up- and down-ramps which the recurrent models exaggerate or flatten. Scheme differences stay smaller than model differences; the point forecasts capture timing but distort peak height, and the integrated variants are the steadiest.}
\begin{table}[!htbp]
\centering
\caption{Peak timing and magnitude errors by model and method. $\Delta h$ is the peak-hour error; $\Delta P_{\text{peak2peak}}$ is the predicted-minus-actual magnitude at each peak; $\Delta P_{\text{at actual}}$ is the error at the actual peak hour. \colorbox{goodcell}{Teal} = good ($|\Delta h|\leq 1$ or $|\Delta P|<200$\,W); \colorbox{badcell}{Red} = poor ($|\Delta h|\geq 4$ or $|\Delta P|>1000$\,W); uncolored = intermediate.}
\label{tab:peak_deltas_grouped}
\begin{tabular}{l l r r r}
\toprule
Model & Day & $\Delta h$ (h) & $\Delta P_{\text{peak2peak}}$ (W) & $\Delta P_{\text{at actual}}$ (W) \\
\midrule
\multicolumn{5}{c}{{Point}} \\
\midrule
\multirow{2}{*}{BiLSTM}     & Jan\,10 & \cellcolor{goodcell}{0} & 502.1 & 502.1 \\
                            & Jan\,12 & \cellcolor{badcell}{6} & -705.6 & \cellcolor{badcell}{-2555.7} \\
\cmidrule{1-5}
\multirow{2}{*}{BiGRU-LSTM} & Jan\,10 & \cellcolor{goodcell}{0} & \cellcolor{goodcell}{13.4} & \cellcolor{goodcell}{13.4} \\
                            & Jan\,12 & \cellcolor{badcell}{5} & \cellcolor{badcell}{-1111.4} & \cellcolor{badcell}{-3423.6} \\
\cmidrule{1-5}
\multirow{2}{*}{TFT}        & Jan\,10 & \cellcolor{goodcell}{0} & \cellcolor{goodcell}{-151.3} & \cellcolor{goodcell}{-151.3} \\
                            & Jan\,12 & \cellcolor{badcell}{5} & \cellcolor{goodcell}{-153.5} & \cellcolor{badcell}{-1267.2} \\
\midrule
\multicolumn{5}{c}{{Modular}} \\
\midrule
\multirow{2}{*}{BiLSTM}     & Jan\,10 & \cellcolor{goodcell}{-1} & \cellcolor{badcell}{2002.0} & \cellcolor{badcell}{1320.9} \\
                            & Jan\,12 & \cellcolor{badcell}{4} & \cellcolor{goodcell}{140.2} & \cellcolor{badcell}{-1280.0} \\
\cmidrule{1-5}
\multirow{2}{*}{BiGRU-LSTM} & Jan\,10 & \cellcolor{badcell}{-4} & \cellcolor{badcell}{1065.1} & 709.6 \\
                            & Jan\,12 & 3 & -862.7 & \cellcolor{badcell}{-2661.1} \\
\cmidrule{1-5}
\multirow{2}{*}{TFT}        & Jan\,10 & \cellcolor{goodcell}{0} & \cellcolor{goodcell}{120.1} & \cellcolor{goodcell}{120.1} \\
                            & Jan\,12 & \cellcolor{badcell}{5} & \cellcolor{goodcell}{90.8} & \cellcolor{badcell}{-1059.8} \\
\midrule
\multicolumn{5}{c}{{Integrated}} \\
\midrule
\multirow{2}{*}{BiLSTM}     & Jan\,10 & \cellcolor{goodcell}{1} & 436.2 & 352.5 \\
                            & Jan\,12 & \cellcolor{badcell}{6} & -673.5 & \cellcolor{badcell}{-3221.0} \\
\cmidrule{1-5}
\multirow{2}{*}{BiGRU-LSTM} & Jan\,10 & \cellcolor{goodcell}{1} & 288.4 & \cellcolor{goodcell}{175.8} \\
                            & Jan\,12 & \cellcolor{badcell}{6} & -842.9 & \cellcolor{badcell}{-3013.2} \\
\cmidrule{1-5}
\multirow{2}{*}{TFT}        & Jan\,10 & \cellcolor{goodcell}{0} & \cellcolor{goodcell}{-14.3} & \cellcolor{goodcell}{-14.3} \\
                            & Jan\,12 & \cellcolor{badcell}{4} & \cellcolor{goodcell}{-12.1} & -291.9 \\
\bottomrule
\end{tabular}
\end{table}

\begin{figure}[!htbp]
  \centering
  \includegraphics[width=\columnwidth]{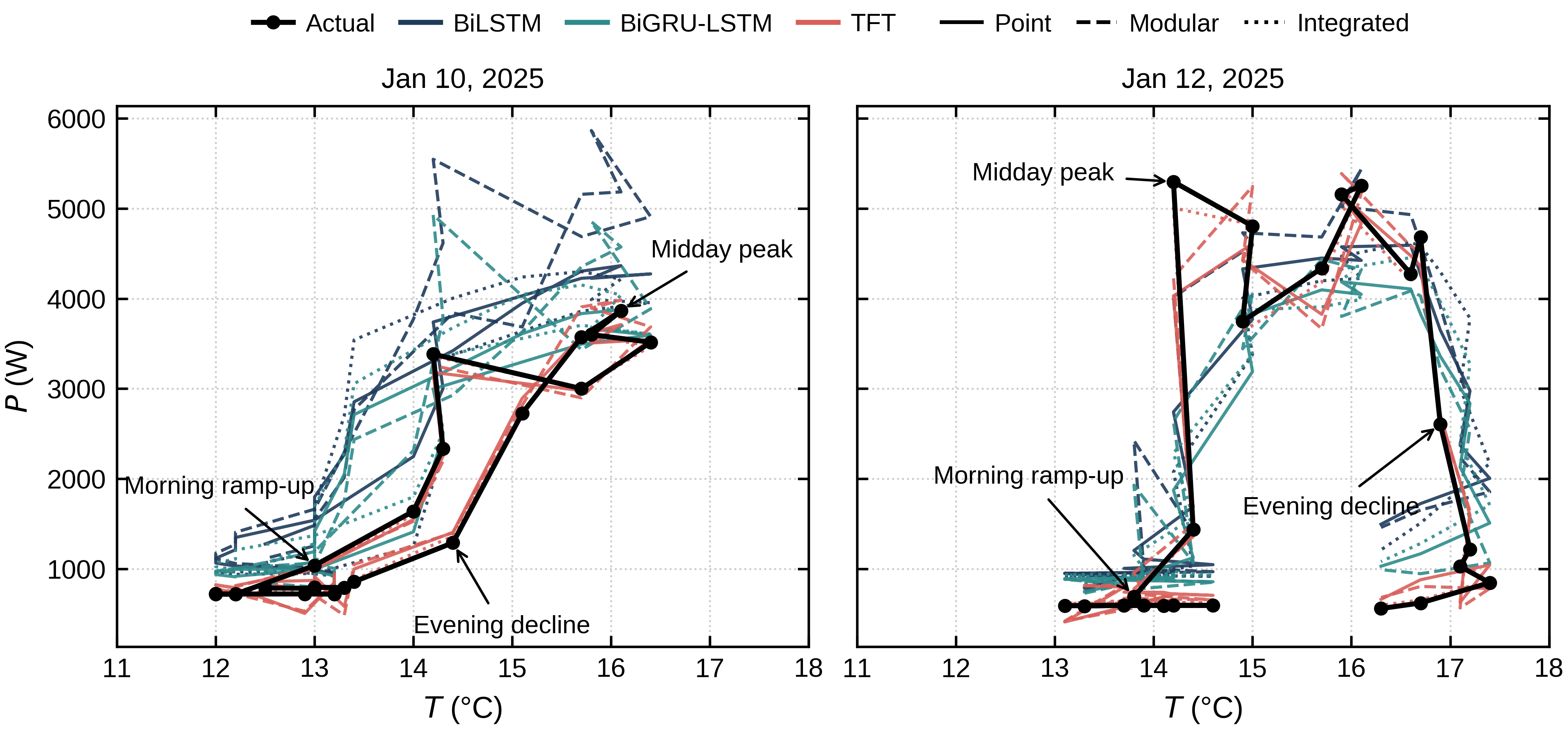}
  \caption{Demand-temperature phase loops for Jan\,10 and 12, 2025, illustrating the clockwise loop.}
  \label{fig:phase_loops}
\end{figure}

\paragraph{Demand-Temperature Loop Tracking}
\textcolor{black}{A final operational concern is whether the forecasts preserve the temperature-demand relationship. Demand typically lags temperature, forming a clockwise daily loop in the demand-temperature plane: morning and evening loads follow different paths even at similar $T$. Fig.~\ref{fig:phase_loops} shows this behavior for the moderate and stress days. On Jan\,12, the stress-day loop is about $6.5\times$ larger than the moderate-day loop (10,729 versus 1,642\,W$\cdot{}^\circ$C), indicating a stronger, more asymmetric load response. The integrated TFT tracks the observed loop best, reducing the mean loop deviation\footnote{\textcolor{black}{Loop deviation is the mean absolute difference between the predicted and observed demand-temperature curves over the 24\,h points of each day, summed over the two days.}} by 79.0\% relative to the point forecast and 79.3\% relative to the modular variant, and best matching the morning and evening branches on the moderate day. The one exception is the stress-day turning-point $T$, where it is slightly worse ($1.9^\circ$C versus $1.7^\circ$C), a small trade-off against its loop-tracking gains.}

\subsubsection{Robustness Analysis}
\paragraph{Comparison with Baselines}
\textcolor{black}{We further test whether the gain of the integrated TFT is due merely to using a deep model. To this end, we compare it with GBM, RF, BPNN/MLP, and SARIMAX under the same inputs, reconstruction procedure, data splits, and horizon $H$. Table~\ref{tab:baselines} shows the non-deep baselines score 43-61\% MAPE, close to the recurrent deep models at 44-70\%, but far from the integrated TFT at 2.7\%. Depth alone thus does not explain the improvement. We attribute the gap to how the architectures handle the reconstructed variables, whose strong training-time correlation with $P$ weakens at test time; the baselines weight them as fixed inputs, whereas the TFT's variable-selection network can re-weight them adaptively. Although the integrated TFT is more robust than the baselines, the following sensitivity analysis isolates the residual cost of reconstruction.}
\begin{table}[!htbp]
\centering
\caption{\textcolor{black}{Statistical and ML baselines vs.\ the integrated TFT on the labeled days (Jan\,10-12) under identical inputs, reconstruction, and splits.}}
\label{tab:baselines}
\begin{tabular}{lccccc}\toprule
Model & RMSE$\downarrow$ & MAE$\downarrow$ & MAPE$\downarrow$ & Accuracy$\uparrow$ & Train/Infer time (s)\\
\midrule
SARIMAX & \textcolor{black}{769.0} & \textcolor{black}{653.2} & \textcolor{black}{61.0} & \textcolor{black}{86.2} & \textcolor{black}{21 / 0.011} \\
GBM & \textcolor{black}{744.1} & \textcolor{black}{631.0} & \textcolor{black}{57.1} & \textcolor{black}{86.7} & \textcolor{black}{1.8 / 0.001} \\
RF & \textcolor{black}{760.4} & \textcolor{black}{649.3} & \textcolor{black}{59.6} & \textcolor{black}{86.3} & \textcolor{black}{18 / 0.026} \\
BPNN/MLP & \textcolor{black}{678.5} & \textcolor{black}{530.0} & \textcolor{black}{43.0} & \textcolor{black}{88.8} & \textcolor{black}{22 / 0.001} \\
TFT (integrated) & \textcolor{black}{48.4} & \textcolor{black}{35.7} & \textcolor{black}{2.7} & \textcolor{black}{99.0} & \textcolor{black}{2,040 / 0.41} \\
\bottomrule
\end{tabular}
\end{table}

\paragraph{Reconstruction Sensitivity}
\textcolor{black}{To isolate the residual cost of reconstruction, we re-evaluate the integrated TFT on $d_2$ using observed versus reconstructed $\{V,I,P^{PV}\}$ channels. Results are pooled over one-day-ahead rollouts on the last fourteen validation days. As shown in Table~\ref{tab:sensitivity}, replacing observed channels with reconstructed ones nearly doubles the QS ($16.5\to34.0$\,W) and lowers coverage from $0.41$ to $0.17$, while the interval width remains almost unchanged ($66.8\to65.5$\,W). Hence, the model does not compensate for reconstruction error by widening its intervals; instead, the predictive distribution shifts, increasing the number of misses. Reconstruction is therefore a major measured source of degradation under feature asymmetry, affecting both sharpness and calibration. Nevertheless, under the same reconstructed inputs, the integrated TFT remains well ahead of the baselines.}
\begin{table}[!htbp]
\centering
\caption{\textcolor{black}{Sensitivity to reconstructed inputs on $d_2$ using the integrated TFT. Reconstructed inputs worsen QS and coverage while leaving interval width nearly unchanged.}}
\label{tab:sensitivity}
\begin{tabular}{lccc}
\toprule
Metric & Observed & Reconstructed & Change \\
\midrule
$\mathrm{QS}$ (W) & 16.5 & 34.0 & $+106\%$ \\
$\mathrm{PICP}_{80}$ & 0.41 & 0.17 & $-0.24$ \\
$\mathrm{MPIW}_{80}$ (W) & 66.8 & 65.5 & $-1.9\%$ \\
\bottomrule
\end{tabular}
\end{table}

\paragraph{Seasonal Robustness}
The labeled competition window is limited to winter, so we test whether the performance is season-specific. We hold out one week each from Apr, Jul, and Oct 2024 from $d_1$, apply identical preprocessing, and re-evaluate the integrated TFT. In Table~\ref{tab:seasonal}, MAPE stays within 1.4-2.9\% across the held-out weeks, consistent with the winter window at 2.7\%. Interval behavior is also stable; coverage meets or exceeds the nominal 80\% (Apr 100\%, Jul 81\%, Oct 93\%) at low $\mathrm{MIS}_{80}$, so the intervals widen on unseen seasons rather than collapse, while the in-distribution winter week is sharper but under-covered (67\%). The good performance of the integrated TFT is thus not confined to the winter competition period. \textcolor{black}{One limitation of this test is that it changes the evaluation season while keeping the training data fixed at the full year, so it does not isolate the effect of the training window on its own. The accuracy and coverage nonetheless stay stable across the held-out weeks, which indicates that full-year training generalizes beyond the winter window rather than overfitting to it. Whether a recent or season-matched training window would give comparable calibration is left to future work.}

\begin{table}[!htbp]
\centering
\caption{\textcolor{black}{Integrated TFT seasonal evaluation. One held-out week per season versus the winter competition window, under identical preprocessing.}}
\label{tab:seasonal}
\begin{tabular}{lccc}\toprule
Season (week) & MAPE$\downarrow$ (\%) & $\mathrm{PICP}_{80}$ (\%) & $\mathrm{MIS}_{80}\downarrow$ (W) \\
\midrule
Apr\,2024 & \textcolor{black}{1.4} & \textcolor{black}{100.0} & \textcolor{black}{260.2} \\
Jul\,2024 & \textcolor{black}{2.9} & \textcolor{black}{81.0} & \textcolor{black}{244.7} \\
Oct\,2024 & \textcolor{black}{1.6} & \textcolor{black}{92.9} & \textcolor{black}{199.5} \\
Jan\,2025 & \textcolor{black}{2.7} & \textcolor{black}{66.7} & \textcolor{black}{175.7} \\
\bottomrule
\end{tabular}
\end{table}

\subsubsection{Remarks and Implications}
\textcolor{black}{Under feature-asymmetric deployment, the uncertainty design matters as much as the backbone. The point forecasts keep timing but develop amplitude bias under stress, and the modular augmentation corrects little, yielding wide, weakly covering intervals and inflated peaks (Table~\ref{tab:prob_all}). The integrated scheme improves calibration without degrading point accuracy (Figs.~\ref{fig:residual_violin} and~\ref{fig:phase_loops}). The backbone ranking is equally clear; TFT attains the lowest errors, the most reliable intervals, and the closest daily patterns (Table~\ref{tab:point_all}; Fig.~\ref{fig:daily-grid-15}). Its edge is suitability to feature asymmetry, not depth. The baselines stay at 43-61\% MAPE under the same setup (Table~\ref{tab:baselines}), and accuracy and calibration hold across held-out seasons (Table~\ref{tab:seasonal}). The remaining bottleneck is the reconstruction itself, which degrades both sharpness and coverage (Table~\ref{tab:sensitivity}). Reliable deployment thus requires integrated probabilistic training with an asymmetry-tolerant backbone such as the TFT, paired with better inference-time reconstruction; the modular augmentation serves better as a diagnostic than an operational scheme.}

\section{Conclusion and Future Work}\label{conc}
Reliable uncertainty estimates in short-term load forecasting affect reserve allocation and demand-response scheduling, yet they are difficult to obtain when deployment inputs are sparse and feature-limited. In the setting studied here, training data are dense and multivariate, whereas inference-time inputs must first be reconstructed from the limited variables available at deployment. Using a three-stage pipeline for data ingestion, temporal alignment, feature reconstruction, and causal feature engineering, we compared two uncertainty strategies, a modular residual-quantile add-on and integrated quantile training, across three deep models: BiLSTM, BiGRU-LSTM, and TFT. The results show that uncertainty placement matters as much as the forecasting backbone. The modular scheme leaves the recurrent backbones with wide, mis-centered intervals and undercoverage as low as 12.5\% on the labeled days. Integrated quantile training narrows the intervals for all three backbones, with the largest gain for the TFT. The integrated TFT gives the lowest errors (RMSE 28-83\,W, MAE 24-56\,W, MAPE 2.2-3.6\%, and accuracy near 99\%), the narrowest intervals, the lowest interval scores, and the coverage closest to nominal at about 67\%. Direct quantile learning is therefore more reliable than post-hoc uncertainty augmentation under the feature-asymmetric deployment studied here, particularly when coupled with an attention-based backbone.

Two limitations define the scope of our findings. First, reconstruction error remains the main technical constraint. Although the reconstructed features preserve the expected input structure, they do not add measured information beyond the available calendar variables and ambient temperature $T$. The true versus reconstructed sensitivity test shows that relying on reconstructed inputs raises the QS by 106\% and drives the 80\% coverage below nominal while leaving the interval width nearly unchanged, so the model does not automatically widen its intervals to reflect reconstruction-induced uncertainty. \textcolor{black}{Second, as our quantitative evaluation rests on one building and 72 labeled hourly observations across three days, the reported rankings should be read as evidence for this deployment setting rather than as population-level conclusions across buildings; we support the rankings with Diebold-Mariano tests under a small-sample correction and with consistent behavior on held-out weeks from other seasons. Future work follows two directions, each addressing one of the limitations above. The first targets the reconstruction-induced miscalibration; uncertainty-aware reconstruction will be coupled with the forecaster so that errors introduced before prediction propagate into the forecast intervals, and lightweight pre-trained sequence models will be explored to build richer representations from the few variables available under sparse sensing. The second broadens evaluation from a single offline deployment toward operational use; the framework will be tested across additional buildings, longer labeled windows, and varying training periods to establish robustness beyond the present setting, and the calibrated forecasts will then drive adaptive control and scheduling through RL-based decision policies.}

\bibliographystyle{elsarticle-harv}
\bibliography{references}
\end{document}